\documentclass[twoside]{article}
\usepackage[T1]{fontenc}
\usepackage[accepted]{aistats2023}
\usepackage{shortcuts}
\usepackage{graphicx, subfigure}
\usepackage{amsmath, amsthm, amstext, amsfonts, amssymb, mathrsfs}
\usepackage{nicefrac}
\usepackage{bbm}
\usepackage[colorlinks=true, urlcolor= black, linkcolor=blue,citecolor=black,pagebackref=true]{hyperref}
\usepackage[round]{natbib}

\usepackage[capitalise]{cleveref}

\newtheorem{lemma}{Lemma}

\newtheorem{proposition}{Proposition}
\newtheorem{theorem}{Theorem}

\newtheorem{corollary}{Corollary}
\theoremstyle{remark} 
\newtheorem{remark}{Remark} 

\usepackage{xcolor}

\colorlet{lightgray}{gray!40}
\usepackage{booktabs}
\usepackage{marvosym}
\usepackage{hhline}
\usepackage{multirow}
\usepackage{colortbl}
\usepackage{algorithm, algorithmic}

\begin{document}

%
\runningtitle{Improved Sample Complexity Bounds For Distributionally Robust Reinforcement Learning}

%

\twocolumn[

\aistatstitle{Improved Sample Complexity Bounds for\\  Distributionally Robust Reinforcement Learning}

\aistatsauthor{ Zaiyan Xu$^*$  \And Kishan Panaganti$^*$  \And Dileep Kalathil }

\aistatsaddress{ Texas A\&M University \And Texas A\&M University \And Texas A\&M University } ]

\begin{abstract}
We consider the problem of learning a control policy  that is robust against the parameter mismatches between the training environment and testing environment. We formulate this as a distributionally robust  reinforcement learning (DR-RL) problem where the objective is to learn the policy which maximizes the value function against the worst possible stochastic model of the environment in an uncertainty set. We focus on the tabular episodic learning setting where the algorithm has access to a generative model of the nominal (training) environment around which the uncertainty set is defined. We propose the Robust Phased Value Learning (RPVL) algorithm to solve this problem for the uncertainty sets specified by four different divergences: total variation, chi-square, Kullback-Leibler, and Wasserstein. We show that our algorithm achieves $\tilde{\cO}(\cardS \cardA H^{5})$   sample complexity, which is uniformly better than the existing results by a factor of $\cardS$, where $\cardS$ is number of states,  $\cardA$ is the number of  actions, and  $H$ is the horizon length. We also provide the first-ever sample complexity result for the Wasserstein uncertainty set. Finally, we demonstrate the performance of our algorithm using simulation experiments. 
\end{abstract}

\section{Introduction}

Training a Reinforcement Learning (RL) algorithm directly on the real-world system is expensive and possibly dangerous since the standard RL algorithms  need a lot of data samples to learn even a reasonably performing policy. The traditional solution to this issue is to train the RL algorithm on a simulator before applying the trained policy to the real-world system. However, there will inevitably be mismatches between the simulator model and the real-world system owing to a variety of factors, including approximation errors made during modeling, changes in the real-world parameters over time, and potential adversarial disruptions in the real-world. For instance, the mass, friction, sensor noise, floor terrain, and obstacles parameters of a mobile robot simulator environment can be different from that of  the real-world environment. Standard  RL algorithms  often fail to perform well when faced with even small changes between the training and testing environments (sim-to-real gap)  \citep{sunderhauf2018limits, tobin2017domain, peng2018sim}.

\begin{table*}[t]
	\begin{center}
\begin{tabular}{|l|c|c|c|c|c|}
\hline
  \multirow{2}{*}{Algorithm}  &  \multicolumn{5}{c|}{Sample Complexity}  \\ \cline{2-6}
   & TV & chi-square & \multicolumn{2}{c|}{Kullback-Leibler} & Wasserstein \\ \hline
\citep{yang2021towards}  & $\frac{\cardS^2\cardA H^5}{\rho^2 \epsilon^2}$ & $\frac{(1+\rho)^2\cardS^2\cardA H^5}{(\sqrt{1+\rho}-1)^2 \epsilon^2}$ & - & $\frac{\cardS^2\cardA H^5}{\rho^2 \pmin^2 \epsilon^2}$ & - \\ \hline
\citep{zhou2021finite}  & - & - & $\frac{\exp{\cO(H)}\cardS^2\cardA H^5}{\rho^2 \epsilon^2}$ & - & - \\ \hline
\citep{shi2022distributionally}  & - & - & - & $\frac{\cardS\cardA H^5}{\rho^2 \pmin \epsilon^2}$ & - \\ \hline
\citep{pmlr-v151-panaganti22a}  & $\frac{\cardS^2\cardA H^5}{ \epsilon^2}$ & $\frac{\rho\cardS^2\cardA H^5}{ \epsilon^2}$ & $\frac{\exp{\cO(H)}\cardS^2\cardA H^5}{\rho^2 \epsilon^2}$ & - & - \\ \hline

\cellcolor{lightgray}\textbf{This work} & $\frac{\cardS\cardA H^5}{\epsilon^2}$ & $\frac{(1+\rho)^2\cardS\cardA H^5}{(\sqrt{1+\rho}-1)^2 \epsilon^2}$ & $\frac{\exp{\cO(H)}\cardS\cardA H^5}{\rho^2 \epsilon^2}$ & $\frac{\cardS\cardA H^5}{\rho^2 \pmin^2 \epsilon^2}$ & $\frac{(B_p+\rho^p)^2\cardS\cardA H^5}{\rho^{2p} \epsilon^2}$ \\ \hline
(Non-robust) Lower bound& \multicolumn{5}{c|}{\,}
\\\citep{li2020breaking} & \multicolumn{5}{c|}{{\centering ${\cardS\cardA H^4}/{\epsilon^2}$}} \\ 
\toprule
\end{tabular}
\end{center}
\caption{ Sample complexity result comparison between our work and the existing works for different uncertainty sets.  Here, $\cardS$ is number of states,  $\cardA$ is the number of  actions,   $H$ is the horizon length, $\rho$ is the uncertainty parameter, $\epsilon\in(0,H)$ is the sub-optimality level,  $\pmin$ is the smallest positive state transition probability of the nominal model, $p\in[1,\infty)$ is the  Wasserstein distance parameter, and $B_p$ is the maximal value of the Wasserstein distance. Results in the infinite horizon setting are adapted to finite horizon setting by using $H$ instead of $(1-\gamma)^{-1}$, where $\gamma$ is the discount factor, and by adding an additional $H$ to accommodate the non-stationary nature of the nominal model. \textit{The sample complexity result of our work is uniformly better than all the existing works by a factor of $\cardS$.  Our bound also matches with the non-robust RL lower bound in all factors, except in $H$.} }
\label{table:comparison-table}
\end{table*}

Learning a policy that is robust against the model parameter mismatches between the training and testing environments is the goal of distributionally robust reinforcement learning (DR-RL). The framework  of the robust Markov decision process (RMDP) \citep{Iyengar-robustDP,nilim-ghaoui-2005} is used to address the distributionally robust \textit{planning} problem. The RMDP formulation considers a collection of models known as the \textit{uncertainty set}, as opposed to the standard non-robust MDP which considers only one model. In DR-RL, the objective is to learn the optimal  robust policy that performs well even under the  worst model in the uncertainty set. The minimization over the uncertainty set makes the RMDP and DR-RL problems significantly more challenging than their non-robust counterparts.

The RMDP problem has been well studied in the literature \citep{xu2010distributionally, wiesemann2013robust, yu2015distributionally, mannor2016robust, russel2019beyond}, taking into account various types of uncertainty sets and computationally effective techniques. These studies, however, are only applicable to the  planning problem. 
Learning algorithms have also been proposed to solve the DR-RL problem \citep{roy2017reinforcement,panaganti2020robust,wang2021online}, but they only provide asymptotic convergence guarantees. There are empirical works that address DR-RL problem using deep RL methods \citep{pinto2017robust, derman2018soft, derman2020bayesian, Mankowitz2020Robust, zhang2020robust}. However, these works do not provide any provable guarantees on the performance of the learned policy.    Recently, there has also been  works on using function approximation approaches with offline data for DR-RL \citep{panaganti-rfqi}.

In this work, we develop a new model-based DR-RL algorithm, Robust Phased Value Learning (RPVL) algorithm, with \textit{provable finite-sample performance guarantees in the tabular, finite-horizon RMDP setting}.
We make the standard generative model assumption used in the model-based non-robust RL literature \citep{AzarMK13, haskell2016empirical, agarwal2020model,li2020breaking, kalathil2021empirical}. More precisely, we assume that the algorithm has  access to a generative  model that can generate next-state samples for all state-action pairs according to the nominal model.
We address the following important (sample complexity) question: \textit{How many samples from the nominal model are required to learn an $\epsilon$-optimal robust policy with a high probability?}

The closest to our work in the DR-RL literature are \cite{yang2021towards,zhou2021finite,pmlr-v151-panaganti22a}. All of these works consider the infinite horizon setting of the DR-RL problem where the nominal model is stationary and obtains a sample complexity $\tilde{\cO}(\cardS^{2})$.    However, many real-world application have non-stationary dynamics and are episodic in nature \citep{choi2009reinforcement,schulman2013finding,zhang2020robust}. So, in this work, we consider a non-stationary nominal model in a finite horizon (episodic) DR-RL setting. The sample complexity of our algorithm is $\tilde{\cO}(\cardS \cardA H^{5})$, which is superior by a factor $\cardS$  compared to all the above mentioned  existing works. \cref{table:comparison-table} provides a detailed comparison of our result with that of the existing works. {We note that a concurrent  work \citep{shi2022distributionally} also considered the finite horizon DR-RL problem but in an offline RL setting and only with Kullback-Leibler uncertainty set. Their sample complexity is $\tilde{\cO}(\cardS)$ which is similar to ours.} 


\paragraph{Our Contributions:}
$(1)$ We propose a new model-based DR-RL algorithm called RPVL algorithm inspired by the \textit{phased value iteration} \citep[Algorithm 4]{kakade-thesis} that takes advantage of the non-stationary dynamics in each \textit{phase}. In addition, we develop an uncertainty-set-specific covering number argument instead of the uniform covering number argument used in the prior works. Combining these, we are able to establish the sample complexity $\tilde{\cO}(\cardS \cardA H^{5})$ for our RPVL algorithm. This not only improves the existing results but also matches with that of the non-robust RL lower bound in $\cardS$ and  $\cardA$  (see \cref{table:comparison-table}). \\
$(2)$ To the best of our knowledge, we provide the  first-ever sample complexity result  for the DR-RL problem with the Wasserstein uncertainty set (see \cref{table:comparison-table}). \\
$(3)$ We demonstrate the performance of our RPVL algorithm on the Gambler's Problem for four different uncertainty sets.  We demonstrate that the learned RPVL algorithm policy is robust to changes in the model parameters for all uncertainty sets. We also show our algorithm converges with regard to the sample size. Finally, with respect to the optimality gap, we demonstrate its dependence on the uncertainty set radius parameter.

\begin{remark}
$(i)$ The term robust RL is now broadly used in a wide variety of formulations, including in the data corruption setting \citep{lykouris2021corruption} and in  the adversarial learning setting \citep{pinto2017robust,zhang2020stability,vinitsky2020robust,zhang2020robust}. The problem addressed in our work is fundamentally different from all these, where we use the classical RMDP framework (Iyengar, 2005; Nilim and El Ghaoui, 2005), and compare with the existing works only using  this framework. $(ii)$ Distributionally robust optimization is now a well established area \citep{duchi2018learning, chen2020distributionally, namkoong2016stochastic}, whose formulation is identical to the classical RMDP formulation. A growing number works \citep{zhou2021finite, si2020distributionally, shi2022distributionally} in the RL literature hence use the terminology  \textit{distributionally} robust to clearly establish this connection and also to avoid possible misinterpretation with the other broad use of the term robust RL. For the same reasons, we use the terminology \textit{DR-RL}  instead of robust RL.
\end{remark}

\section{Preliminaries and Problem Formulation}

\paragraph{Notations:}$\Delta(\states)$ denotes the probability simplex over any finite set $\states$. For any probability vector $P\in\Delta(\states)$ and real vector $V\in\bR^{\cardS}$, let $PV$ denote the expectation $\expectsim{s\sim P}{V(s)}$. {For any positive integer $M$, $[M]$ denotes the set $\{1, 2, \ldots, M \}$}. $\vecofone$ denotes the vector of all ones whose dimension is determined from context. Let $\mathrm{m}(\mu,\nu)\subseteq \Delta(\states\times\states)$ denote the set of all probability measures on $\states\times\states$ such that the marginals along first and second dimensions are the distribution $\nu$ and $\mu$ respectively. That is, for any $\omega\in\mathrm{m}(\mu,\nu)$ we have $\nu(y)=\int_x \omega(x,y) dx$ and $\mu(x)=\int_y \omega(x,y) dy$.

A \textbf{finite-horizon Markov Decision Process (MDP)} can be defined as a tuple $(\states,\actions,(P_h)_{h=1}^{H},(r_h)_{h=1}^{H}, {H})$, where $\states$ is the state space, $\actions$ is the action space, $H$ is the horizon length,  for any $h \in [H]$, $r_h\colon\states\times\actions\to[0,1]$ is a known deterministic reward function, and $P_h\colon\states\times\actions\to\Delta(\states)$ is the transition probability function at time $h$. For any $h\in [H]$, and $P_h(s'| s,a)$ represents the probability of transitioning to state $s'$ when action $a$ is taken at state $s$. We use   $P_{h,s,a}$ to  denote the $\cardS$-dimensional probability vector taking value $P_{h,s,a}(s') := P_h(s'| s,a)$ for any $s'\in\states$. We assume that $\cardS$ and $\cardA$ are finite. When it is clear from the context, we  use  the shorthand $P$ for $(P_h)_{h=1}^{H}$. 
 
A non-stationary Markov policy  $\pi=(\pi_h)_{h=1}^H$ is a sequence of decision rules such that  $\pi_h\colon \states\to \Delta(\actions)$ where $\pi_h(a| s)$ specifies the probability of choosing action $a$ in state $s$ at time $h$. We consider the deterministic Markov policy class $\Pi$. That is, for any policy $\pi\in\Pi$, $\pi_h(s)$ is a deterministic decision rule.   For any state $s\in\states$ and time $h\in[H]$, we define the value function for policy $\pi$ as $ V_h^{\pi,P}(s) := \expectsim{\pi,P}{\sum_{t=h}^H r_t(s_t,a_t) \mid s_h = s, \pi}$, where $a_h\sim\pi_h(s_h)$ and $s_{h+1}\sim P_h(\cdot| s_h, a_h)$. Since we are particularly interested in the value of a policy starting from  $h=1$, we denote $V^{\pi,P}:=V_1^{\pi,P}$. The optimal  value function $V^{*,P}_{h}$ and the  optimal policy  $\pi^{*,P}_{h}$ of an MDP with the transition dynamics $P$ are defined as
\begin{align*}
    V^{*,P}_{h} = \max_{\pi\in\Pi} V_{h}^{\pi,P},~ \pi^{*,P}_{h} = \argmax_{\pi\in\Pi} V_{h}^{\pi,P},~\forall h\in[H]. 
\end{align*}
Since we assume that $r_h(s,a)\in[0,1]$ for all $(h,s,a)\in[H]\times\states\times\actions$, it is straightforward that any value function is bounded by $V_{\max}:=H$. We denote the set of all value functions $\cV=\curlyns{V\in\bR^{\cardS}\colon 0\leq \supnormns{V} \leq H}$.

\paragraph{(Distributionally) Robust Markov Decision Process:}
A finite-horizon distributionally robust MDP (RMDP) can be defined as a tuple $M = (\states, \actions, \cP, (r_h)_{h=1}^{H}, H)$. At each time $h\in[H]$, instead of a single model, we consider the set of models within  a ball centered at the nominal model $P^o\equiv (P^o_h)_{h=1}^H$. Formally, we define the uncertainty set as $\cP = \bigotimes_{h,s,a\in[H]\times\states\times\actions} \cP_{h,s,a}$ such that
\begin{equation}\label{eq:uncertainty-set-def}
    \cP_{h,s,a} = \curly{P\in\Delta(\states) \colon D(P,P^o_{h,s,a}) \leq \rho},
\end{equation}
where $D(\cdot,\cdot)$ is some distance metric between two probability measures, and $\rho$ defines the radius of the uncertainty set. We note that, by construction,  the uncertainty set $\cP$  satisfies $(s,a)$-rectangularity condition \citep{Iyengar-robustDP}.

\paragraph{Choices of Uncertainty Sets:} We focus on four different uncertainty sets - three corresponding to different $f$-divergences and one corresponding to Wasserstein metric.

$(i)$ \textit{Total variation uncertainty set:} Let $\cP^{\mathrm{TV}}=\bigotimes_{h,s,a}\cP_{h,s,a}^{\mathrm{TV}}$, where $\cP_{h,s,a}^{\mathrm{TV}}$ is defined as in \cref{eq:uncertainty-set-def} with total variation distance
    \begin{equation}
        D_{\mathrm{TV}}(P,P^o_{h,s,a}) = (1/2)\onenorm{P-P^o_{h,s,a}}. \label{eq:D-TV}
    \end{equation}
$(ii)$ \textit{Chi-square uncertainty set:} Let $\cP^\chi=\bigotimes_{h,s,a}\cP_{h,s,a}^{\chi}$, where $\cP_{h,s,a}^\chi$ is defined as in \cref{eq:uncertainty-set-def} with Chi-square distance
    \begin{equation} \label{eq:D-chi}
        D_\chi(P,P^o_{h,s,a}) = \sum_{s'\in\states}\frac{(P(s')-P^o_{h,s,a}(s'))^2}{P^o_{h,s,a}(s')}.
    \end{equation}
$(iii)$ \textit{Kullback-Leibler uncertainty set:} Let $\cP^{\mathrm{KL}}=\bigotimes_{h,s,a}\cP_{h,s,a}^{\mathrm{KL}}$, where $\cP^{\mathrm{KL}}_{h,s,a}$ is defined as in \cref{eq:uncertainty-set-def} with Kullback-Leibler distance
    \begin{equation} \label{eq:D-KL}
        D_{\mathrm{KL}}(P,P^o_{h,s,a}) = \sum_{s'\in\states}P(s')\log{\frac{P(s')}{P^o_{h,s,a}(s')}}.
    \end{equation}
$(iv)$ \textit{Wasserstein uncertainty set:} Let $\cP^{\mathrm{W}}=\bigotimes_{h,s,a}\cP_{h,s,a}^{\mathrm{W}}$, where $\cP^{\mathrm{W}}_{h,s,a}$ is defined as in \cref{eq:uncertainty-set-def} with Wasserstein distance
    \begin{equation} \label{eq:D-wass}
        D_{\mathrm{W}}(P,P^o_{h,s,a}) = \inf_{\nu\in \mathrm{m}(P,P^o_{h,s,a})} \int d^p(x,y)d\nu(x,y),
    \end{equation}
    where the integration is over $(x,y)\in\states\times\states$, $p\in[1,\infty)$, and $\mathrm{m}(P,P^o_{h,s,a})$ denotes all probability measures on $\states\times\states$ with marginals $P$ and $P^o_{h,s,a}$. In addition, we set $B_p:=\max_{s,s'}d^p(s,s')$.

\paragraph{Robust Dynamic Programming:} Under the RMDP setting, for a policy $\pi\in\Pi$ the robust value function $V^\pi$ is defined as \citep{Iyengar-robustDP,nilim-ghaoui-2005}
\begin{align}
    V_h^\pi(s) &= \inf_{P\in\cP} V_h^{\pi,P}(s), ~~ \forall (h,s)\in[H]\times\states.
\end{align}
The interpretation is that we  evaluate a policy using the worst possible model in the uncertainty set. For consistency, we set $V_{H+1}^{\pi}=0$ for any $\pi\in\Pi$. 

The  optimal robust value function and the optimal policy are defined as, $\forall (h,s)\in[H]\times\states$, 
\begin{align}
    V^*_h(s) &= \sup_{\pi\in\Pi} V_h^\pi(s),~~ \pi^{*}_{h} = \arg \sup_{\pi\in\Pi} V_h^\pi(s),  \label{eq:opt-robust-value}
\end{align}
It is known that  there exists at least one optimal deterministic Markov policy \citep[Theorem 2.2]{Iyengar-robustDP}.

To simplify the notations,  denote 
\begin{align}
    L_{\cP_{h,s,a}}V = \inf\curlyns{PV\colon P\in\cP_{h,s,a}}.
\end{align}
The optimal robust value function satisfies the following \textit{robust Bellman  equation} \citep[Theorem 2.1]{Iyengar-robustDP}, 
\begin{align}
    V^*_h(s) &= \max_{a\in\actions}\curlyns{r_h(s,a) + L_{\cP_{h,s,a}}V^*_{h+1}}, \label{eq:robust-bellman-opt-eq}
\end{align}
for $\forall (h,s)\in[H]\times\states$. From the  optimal robust value function, the optimal robust policy can be calculated as
\begin{align}
\label{eq:robust-policy-1}
    \pi^*_h(s) &= \argmax_{a\in\actions}\curlyns{r_h(s,a) + L_{\cP_{h,s,a}}V^*_{h+1}},
\end{align}
for $\forall (h,s)\in[H]\times\states$. The \textit{robust dynamic programming} method involves performing the  backward iteration  using \cref{eq:robust-bellman-opt-eq}-\cref{eq:robust-policy-1} to find $V^{*} = (V^{*})^{H}_{h=1}$ and $\pi^{*} = (\pi^{*})^{H}_{h=1}$.

We also have the \textit{robust Bellman consistency equation} \citep[Theorem 2.1]{Iyengar-robustDP} which is the extension of its counterpart in standard MDP. For any $\pi\in\Pi$ and $\forall h\in[H]$,
\begin{align}
    V^\pi_h(s) = r_h(s,\pi_h(s)) + L_{\cP_{h,s,\pi_h(s)}}V^\pi_{h+1},~ \forall s\in\states. \label{eq:robust-consistency-eq}
\end{align}

\section{Algorithm and Sample Complexity}

In order to compute $V^*$ and $\pi^*$, the robust dynamic programming method requires the knowledge of the nominal model $P^o$ and the radius of the uncertainty set $\rho$. Although $\rho$ could be a design parameter, in most practical applications, we do not have access to the analytical form of the nominal model $P^{o}$. So, \textit{we assume that the nominal model $P^o$ is unknown}. Instead, similar to the standard non-robust RL setting, we assume that we only have access to samples from a generative model, which, given the state-action $(s,a)$ and time $h$ as inputs, will produce samples of the next state $s'$ according to $P^o_{h,s,a}$. We propose a model-based algorithm called Robust Phased Value Learning (RPVL) algorithm (\cref{rpvlalgo}) to learn the optimal robust policy.

RPVL algorithm is inspired by the phased value iteration in non-robust RL \cite[Algorithm 4]{kakade-thesis}.
We first get a maximum likelihood estimate $\Phat^{o}_{h}$ of the nominal non-stationary model $P^{o}_{h}$ for each $h\in[H]$ by following the standard approach in the literature  \cite[Algorithm 4]{kakade-thesis} \cite[Algorithm 3]{AzarMK13}. The term \textit{phased}  indicates that we use different independent samples to estimate the nominal non-stationary model $P^{o}_{h}$ for each step (phase) $h\in[H]$. Formally, for  each  $h\in[H]$, we generate $N$ next-state samples corresponding to  each state-action pairs. Let $N_h(s,a,s')$ be the count of the state $s'$ in the $N$ total  transitions from the state-action pair $(s, a)$ in step  $h\in[H]$. Now, the   maximum likelihood estimate  is given by $\widehat{P}^{o}_{h,s,a}(s') = {N_h(s,a,s')}/{N}$. Given $\widehat{P}^{o}$, we can get  an empirical estimate of the uncertainty set $\cP$ as, 
\begin{align}
    \label{eq:uncertainty-set-estimate}
    \cPhat = &\bigotimes_{h,s,a\in[H]\times\states\times\actions} \cPhat_{h,s,a},~\text{where},\\
    &\cPhat_{h,s,a} = \curly{P\in\Delta(\states) \colon D(P,\Phat^o_{h,s,a}) \leq \rho}, \nonumber
\end{align} 
where $D$ is one of the metrics  specified in \cref{eq:D-TV} - \cref{eq:D-wass}.

For finding an approximately optimal robust policy, we
now consider the empirical RMDP $\widehat{M}  = (\states, \actions, \widehat{\cP}, (r_h)_{h=1}^{H}, H)$ and perform robust dynamic programming as given in  \cref{eq:robust-bellman-opt-eq}-\cref{eq:robust-policy-1}. We formally give our proposed approach in \cref{rpvlalgo}. 

\begin{algorithm}[t!]
	\caption{Robust Phased Value Learning (RPVL)}	
	\label{rpvlalgo}
	\begin{algorithmic}[1]
	\STATE \textbf{Input:} Uncertainty radius $\rho$
		\STATE \textbf{Initialize:} $\Vhat_{H+1} = 0$
		\FOR {$h=H,\ldots,1$ }
		\STATE Compute the empirical uncertainty set $\cPhat_{h,s,a}$ for all $(s,a)\in(\states,\actions)$ according to \cref{eq:uncertainty-set-estimate}
		\STATE $\Vhat_{h}(s) = \max_{a} (r(s,a) + L_{\cPhat_{h,s,a}} \Vhat_{h+1}),~\forall s\in\states$
		\STATE $\pihat_{h}(s) = \argmax_{a} (r(s,a) + L_{\cPhat_{h,s,a}} \Vhat_{h+1}),~\forall s\in\states$
		\ENDFOR
		
		\STATE \textbf{Output:} $\pihat = (\pihat_h)_{h=1}^H$
	\end{algorithmic}
\end{algorithm}

\subsection{Sample Complexity Results}

Now we give the sample complexity results for all uncertainty sets based on the metrics  specified in \cref{eq:D-TV} - \cref{eq:D-wass}. We  note that the total number of samples needed in the RPVL algorithm is $N_{\mathrm{total}} = N \cardS \cardA H$.

\begin{theorem}[TV uncertainty set.]
\label{thm:tv-vhat-theorem}
Consider a finite-horizon RMDP with a total variation uncertainty set $\cP^{\mathrm{TV}}$. Fix $\delta\in(0,1)$, $\rho>0$, and $\epsilon\in(0, 8H)$. Consider the RPVL algorithm, with the total number of samples $N_{\mathrm{total}}\geq N_{\mathrm{TV}}$, where
{\small\begin{equation*}
    N_{\mathrm{TV}} = \frac{8 H^5\cardS\cardA}{\epsilon^2}\cdot\log{\frac{32 H^3\cardS\cardA}{\epsilon\delta}}.
\end{equation*}}
Then, $\supnormns{V^*_1 - V^{\pihat}_1} \leq \epsilon$, with probability at least $1-\delta$.
\end{theorem}

\begin{remark}[Comparison with the sample complexity of the non-robust RL]
\label{rem:comarison-with-nonrobust-RL} It is known from the non-robust non-stationary RL literature \citep{AzarMK13} that, for $\epsilon\in(0,1)$, no algorithm can learn an $\epsilon$-optimal policy with fewer than $\tilde{\Omega}({\cardS \cardA H^4}/{\epsilon^{2}})$ samples. Recently,  \cite{li2020breaking} settled this sample complexity question with a matching lower and upper bound $\tilde{\Theta}({\cardS \cardA H^4}/{\epsilon^{2}})$ for the \textit{full} range of accuracy level $\epsilon \in (0, H)$. Our sample complexity bound given in \cref{thm:tv-vhat-theorem} for the distributionally robust RL is $\tilde{\cO}(\cardS \cardA H^{5}/\epsilon^{2})$, which matches with that of the non-robust RL bound in $\cardS$, $\cardA$ and also accommodates the \textit{full} range of accuracy level $(0,H)$, albeit worse by a factor $H$.
\end{remark}

\begin{theorem}[Chi-square uncertainty set]
\label{thm:chi-vhat-theorem}
Consider a finite-horizon RMDP with a chi-square uncertainty set $\cP^{\chi}$. Fix $\delta\in(0,1)$, $\rho>0$, and $\epsilon\in(0, 8H)$. Consider the RPVL algorithm, with the total number of samples $N_{\mathrm{total}}\geq N_{\chi}$, where
{\small\begin{equation*}
    N_{\chi} = \frac{128 C_\rho^4 H^5\cardS\cardA}{(C_\rho - 1)^2\epsilon^2}\cdot\log{\frac{2H\cardS\cardA(1+\frac{8C_\rho H^2}{\epsilon(C_\rho - 1)})}{\delta}},
\end{equation*}}
and where $C_\rho=\sqrt{1+\rho}$. Then, $\supnormns{V^* - V^{\pihat}} \leq \epsilon$ with probability at least $1-\delta$.
\end{theorem}
We note that \cref{rem:comarison-with-nonrobust-RL} holds for the above theorem also.

For KL uncertainty set, we present two results. 
 
\begin{theorem}[KL uncertainty set]
\label{thm:kl-exp-vhat-theorem}
Consider a finite-horizon RMDP with a KL uncertainty set $\cP^{\mathrm{KL}}$. Fix $\delta\in(0,1)$, $\rho>0$, and $\epsilon\in(0, H)$. Consider the RPVL algorithm, with the total number of samples $N_{\mathrm{total}}\geq N_{\mathrm{KL}}$, where
{\small\begin{equation*}
    N_{\mathrm{KL}} = \frac{2\exp{3H/\underlambda} H^5\cardS\cardA}{\rho^2\epsilon^2} \cdot\log{\frac{8H\cardS\cardA}{\underlambda\delta}},
\end{equation*}}
and $\underlambda$ is a problem dependent parameter independent of $N_{\mathrm{KL}}$, then $\supnormns{V^* - V^{\pihat}} \leq \epsilon$ with probability at least $1-\delta$.
\end{theorem}
We note that the sample complexity of the above result for the KL uncertainty set is worse by a term $\exp{3H/\underlambda}$ when compared to \cref{thm:tv-vhat-theorem} and \cref{thm:chi-vhat-theorem}. We note that this exponential dependence on $H$ is observed in the prior works also \citep{pmlr-v151-panaganti22a, zhou2021finite}. We can, however, replace the exponential dependence on horizon with a \textit{problem dependent constant} as stated in the the results below. 
\begin{theorem}[KL uncertainty set]
\label{thm:kl-pmin-vhat-theorem}
Consider a finite-horizon RMDP with a KL uncertainty set $\cP^{\mathrm{KL}}$. Fix $\delta\in(0,1)$, $\rho>0$, and $\epsilon\in(0, H)$. Consider the RPVL algorithm, with the total number of samples $N_{\mathrm{total}}\geq N'_{\mathrm{KL}}$, where
{\small\begin{equation*}
    N'_{\mathrm{KL}} = \frac{2H^5\cardS\cardA}{\rho^2\pmin^2\epsilon^2} \cdot\log{\frac{2H\cardS^2\cardA}{\delta}},
\end{equation*}}
and where $\pmin=\min_{s',s,a,h: P^o_h(s'| s,a) >0} P^o_h(s'| s,a)$. Then, $\supnormns{V^* - V^{\pihat}} \leq \epsilon$ with probability at least $1-\delta$.
\end{theorem}
The sample complexity of the RPVL algorithm with the KL  uncertainty set here is $\tilde{\cO}( {\cardS  \cardA H^5}/{ \rho^2\pmin^2 \epsilon^2})$. We note that $\pmin$ is a problem dependent constant and can be arbitrarily low. So, $1/\pmin^{2}$ can even be larger than $\exp{3H/\underlambda}$. So, this also does not offer a definitive answer to the question of finding a problem-independent sample complexity bound for the KL uncertainty set, which is still an open question. We, however, emphasize that our sample complexity bounds given in \cref{thm:kl-exp-vhat-theorem} and \cref{thm:kl-pmin-vhat-theorem} are superior compared to the existing results  by a factor of $\cardS$ (see \cref{table:comparison-table} for details). In addition, \cref{rem:comarison-with-nonrobust-RL} holds for both \cref{thm:kl-exp-vhat-theorem} and \cref{thm:kl-pmin-vhat-theorem}.

\begin{theorem}[Wasserstein uncertainty set]
\label{thm:wasserstein-theorem}
Consider a finite-horizon RMDP with a Wasserstein uncertainty set $\cP^{\mathrm{W}}$. Fix $\delta\in(0,1)$, $\rho>0$, and $\epsilon\in(0, H)$. Consider the RPVL algorithm, with the total number of samples $N_{\mathrm{total}}\geq N_{\mathrm{W}}$, where
{\small\begin{align*}
    N_{\mathrm{W}} = &\frac{8H^5\cardS\cardA(B_p+\rho^p)^2}{\rho^{2p}\epsilon^2}\cdot\\&\hspace{1cm}\log{\frac{16H^2\cardS\cardA(HB_p+(H\vee\rho^p))}{\rho^p\delta\epsilon}}.
\end{align*}}
Then, $\supnormns{V^* - V^{\pihat}} \leq \epsilon$ with probability at least $1-\delta$.
\end{theorem}
We note that \cref{rem:comarison-with-nonrobust-RL} also holds for the Wasserstein uncertainty set case as 
the sample complexity here is $\tilde{\cO}( { (B_p+\rho^p)^2\cardS  \cardA H^5}/{ \rho^{2p}\epsilon^2})$.

\section{Sample Complexity Analysis}

Here  we briefly explain the key analysis ideas of RPVL algorithm for obtaining the sample complexity in \cref{thm:tv-vhat-theorem}~-~\cref{thm:wasserstein-theorem}. The complete proofs are provided in \cref{appen:tv-vhat-theorem}~-~\cref{appen:wasserstein-theorem}.

\textit{Step 1:} To bound $ \supnormns{V^*_1 - V^{\pihat}_1}$, we split it up using the RPVL algorithm value function $\Vhat_1$ as $\supnormns{V_1^* - V_1^{\pihat}} \leq \supnormns{V_1^* - \Vhat_1} + \supnormns{\Vhat_1 - V_1^{\pihat}}.$ This split is useful since it is easier to bound these individually. We now state a result \citep{yang2021towards,pmlr-v151-panaganti22a,shi2022distributionally}  which shows Lipschitz property of operators $L_{\cP_{h,s,a}}$ and $L_{\cPhat_{h,s,a}}$ for all $h,s,a\in[H]\times\states\times\actions$. 
\begin{lemma}[\text{\citealp[Lemma 1]{pmlr-v151-panaganti22a}}]\label{lem:uncertainty-inf-lipschitz}
For any $h,s,a\in[H]\times\states\times\actions$ and for any $V_1,V_2\in\bR^{\cardS}$, we have $\absns{L_{\cP_{h,s,a}}V_1 - L_{\cP_{h,s,a}}V_2}\leq\supnormns{V_1-V_2}$ and $\absns{L_{\cPhat_{h,s,a}}V_1 - L_{\cPhat_{h,s,a}}V_2}\leq\supnormns{V_1-V_2}$.
\end{lemma}
Using this and the \textit{robust Bellman equations} \cref{eq:robust-bellman-opt-eq} and \cref{eq:robust-consistency-eq} in tandem, we can establish the recursion
\begin{align*}
    \supnormns{V_h^* - \Vhat_h} &\leq \supnormns{V_{h+1}^* - \Vhat_{h+1}^*} \\&\hspace{1cm}+ \max_{h,s,a}\abs{L_{\cP_{h,s,a}}\Vhat_{h+1} - L_{\cPhat_{h,s,a}}\Vhat_{h+1}},\\
    \supnormns{V_h^{\pihat} - \Vhat_h} &\leq \supnormns{V_{h+1}^{\pihat} - \Vhat_{h+1}} \\&\hspace{1cm}+ \max_{h,s,a}\abs{L_{\cP_{h,s,a}}\Vhat_{h+1} - L_{\cPhat_{h,s,a}}\Vhat_{h+1}}.
\end{align*}
Using the consistency of robust value functions at $H+1$, i.e., $V_{H+1}^{\pi}=0$ for any $\pi\in\Pi$ and $\Vhat_{H+1}=0$, from the recursion on $h$, we finally obtain 

\begin{align}\label{eq:main-step-1}
    \supnormns{V_1^* - &V_1^{\pihat}}\qquad\qquad\qquad\qquad\qquad\qquad\qquad\qquad \nonumber\\
    &\leq 2H \max_{h,s,a}\absns{L_{\cP_{h,s,a}}\Vhat_{h+1} - L_{\cPhat_{h,s,a}}\Vhat_{h+1}}.
\end{align}
\textit{Step 2:} This is the crucial step where we develop concentration inequality bounds for $\max_{h,s,a}\absns{L_{\cP_{h,s,a}}\Vhat_{h+1} - L_{\cPhat_{h,s,a}}\Vhat_{h+1}}$ in \cref{eq:main-step-1}. We emphasize that this step is different and depends on the structure of $L_{\cP_{h,s,a}}$ and $L_{\cPhat_{h,s,a}}$ for all uncertainty sets considered in this work.

Obtaining a bound for  $\max_{h,s,a}\absns{L_{\cP_{h,s,a}}\Vhat_{h+1} - L_{\cPhat_{h,s,a}}\Vhat_{h+1}}$ is the most challenging part of our analysis. In the non-robust setting, this  will be equivalent to the error term $P^{o}_{h,s,a} V - \Phat_{h,s,a} V$, which is unbiased and can be easily bounded using  concentration inequalities. In the distributionally robust setting, establishing such a concentration  is not immediate because  the  nonlinear nature of the function $L(\cdot)$ leads to  $\bE[L_{\cPhat_{h,s,a}}\Vhat_{h+1}] \neq L_{\cP_{h,s,a}}\Vhat_{h+1}$. We note that the previous works \citep{panaganti2021sample,yang2021towards,zhou2021finite,pmlr-v151-panaganti22a,shi2022distributionally} also develop such concentration inequalities for the distributionally robust setting. But all of them rely on developing the uniform bounds over the value function space yielding worse sample complexity with an extra $\cardS$ factor.
Instead, our strategy is to get amenable concentration inequalities with novel covering number arguments \cref{prop:robust-stochastic-value-est-error-TV,prop:robust-stochastic-value-est-error-CHI,prop:robust-stochastic-value-est-error-KL-EXP,prop:robust-stochastic-value-est-error-WASSERSTEIN}.
In our analyses, we also make use of an important RPVL algorithm property that $\Vhat_{h+1}$ is independent of $\Phat^o_{h,s,a}$ by construction. We emphasize that this is the first work to use this property in the DR-RL literature. This technique is inspired by \citep[Theorem 2.5.1]{kakade-thesis} in robust RL literature. Below, we outline the analysis of the Step 2  separately for all the uncertainty sets.

\paragraph{Total Variation Uncertainty Set:} We first state a result  which gives an equivalent dual form characterization for the optimization problem $L_{\cP_{h,s,a}}V$, where  $\cP_{h,s,a}$ is the total variation uncertainty set $\cP^{\mathrm{TV}}_{h,s,a}$. 
\begin{proposition}[\text{\citealp[Lemma 5]{panaganti-rfqi} }]\label{prop:inner-dual-tv}
Consider an RMDP $M$ with the total variation uncertainty set $\cP^{\mathrm{TV}}$. Fix any $h,s,a\in[H]\times\states\times\actions$. For any value function $V\in\cV$, the inner optimization problem $L_{\cP^{\mathrm{TV}}_{h,s,a}}V$ has the following equivalent form
\begin{align*}
    &L_{\cP^{\mathrm{TV}}_{h,s,a}}V = - \inf_{\eta\in[0,2 H/\rho]} \expectsim{s'\sim P^o_h(\cdot| s,a)}{(\eta - V(s'))_+} \\&\hspace{3.5cm}+ \big(\eta - \inf_{s''\in\states}V(s'')\big)_+ \cdot \rho - \eta.
\end{align*}
\end{proposition}
\begin{remark} \label{remark:dual-form-advantage}
We remark that the dual form of the optimization problem $L_{\cP^{\mathrm{TV}}_{h,s,a}}V$ above is a single variable convex optimization problem.
This result hence is useful for both the analysis of sample complexity bound and for the practical implementation of our RPVL algorithm.
\end{remark}
Now we develop our main concentration inequality result based on \cref{prop:inner-dual-tv} and the RPVL algorithm property that $\Vhat_{h+1}$ is independent of $\Phat^o_{h,s,a}$ by construction.
\begin{proposition}\label{prop:robust-stochastic-value-est-error-TV}
Fix any $h,s,a\in[H]\times\states\times\actions$. For any $\theta,\delta\in(0,1)$ and $\rho\in(0,1]$, we have, with probability at least $1-\delta$,
$\absns{L_{\cP^{\mathrm{TV}}_{h,s,a}}\Vhat_{h+1} - L_{\cPhat^{\mathrm{TV}}_{h,s,a}}\Vhat_{h+1}} \leq 2\theta +  \sqrt{{H^2\log{{4H}/{(\theta\delta)}}}/{(2N)}}.$

\end{proposition}

\begin{remark}
\label{rem:covering-theta}
We  note  that the $\theta$ here is a covering number argument parameter (see \cref{lem:cover-number-tv}) which is a novel development made in this work which also enabled us to improve the sample complexity. 
\end{remark}
Finally, the proof of \cref{thm:tv-vhat-theorem} is now complete by the application of uniform bound over the $[H]\times\states\times\actions$ space, combined with \cref{eq:main-step-1}.

\paragraph{Chi-square Uncertainty Set:} We first develop the dual reformulation version of $L_{\cP^{\chi}_{h,s,a}}V$ by adapting the techniques from distributionally robust optimization \citep{duchi2018learning}. 
\begin{proposition}\label{prop:inner-dual-chi}
Consider an RMDP $M$ with the chi-square uncertainty set $\cP^\chi$. Fix any $h,s,a\in[H]\times\states\times\actions$. For any value function $V\in\cV$, the inner optimization problem $L_{\cP^{\chi}_{h,s,a}}V$ has the following equivalent form
\begin{align*}
    &L_{\cP^{\chi}_{h,s,a}}V \\&= - \inf_{\eta\in \Xi} \curly{C_\rho \sqrt{\expectsim{s'\sim P^o_{h}(\cdot| s,a)}{(\eta-V(s'))^2}} - \eta},
\end{align*}
where $\Xi = [0,C_\rho H/(C_\rho -1)]$ and $C_\rho=\sqrt{\rho + 1}$.
\end{proposition}
\cref{remark:dual-form-advantage} hold here as well. 
Now we develop our main concentration inequality result based on \cref{prop:inner-dual-chi} and the RPVL algorithm property that $\Vhat_{h+1}$ is independent of $\Phat^o_{h,s,a}$ by construction.
\begin{proposition}\label{prop:robust-stochastic-value-est-error-CHI}
Fix any $h,s,a\in[H]\times\states\times\actions$. For any $\theta,\delta\in(0,1)$ and $\rho>0$, we have, with probability at least $1-\delta$,
\begin{align*}
    &\absns{L_{\cP^{\chi}_{h,s,a}}\Vhat_{h+1} - L_{\cPhat^{\chi}_{h,s,a}}\Vhat_{h+1}} \leq 2\theta + \\& \frac{\sqrt{2}C_\rho^2 H}{(C_\rho - 1)\sqrt{N}} \big(\sqrt{\log{\frac{2(1+C_\rho H/(\theta(C_\rho-1)))}{\delta}}} + 1\big).
\end{align*}
\end{proposition}
\cref{rem:covering-theta} holds here as well (see \cref{lem:cover-with-lp-norm}). 
Finally, the proof of \cref{thm:chi-vhat-theorem} is now complete by the application of uniform bound over the $[H]\times\states\times\actions$ space, combined with \cref{eq:main-step-1}.

\paragraph{KL Uncertainty Set:} Here also, the first step is to formulate the dual version of $L_{\cP^{\mathrm{KL}}_{h,s,a}}V$ as given below.  
\begin{proposition}\label{prop:inner-dual-kl}
Consider an RMDP $M$ with the chi-square uncertainty set $\cP^{\mathrm{KL}}$. Fix any $h,s,a\in[H]\times\states\times\actions$. For any value function $V\in\cV$, the inner optimization problem $L_{\cP^{\mathrm{KL}}_{h,s,a}}V$ has the following equivalent form
\begin{align*}
    &L_{\cP^{\mathrm{KL}}_{h,s,a}}V = - \inf_{\lambda\in [0,H/\rho]} \big(\lambda\rho \\&\hspace{1.5cm}+ \lambda \log{\expectsim{s'\sim P^o_h(\cdot| s,a)}{\exp{\frac{-V(s')}{\lambda}}}}\big).
\end{align*}
\end{proposition}
Now we develop our main concentration inequality result based on \cref{prop:inner-dual-kl} and the RPVL algorithm property that $\Vhat_{h+1}$ is independent of $\Phat^o_{h,s,a}$ by construction.
\begin{proposition}\label{prop:robust-stochastic-value-est-error-KL-EXP}
Fix any $h,s,a\in[H]\times\states\times\actions$. For any $\theta,\delta\in(0,1)$ and $\rho>0$, we have, with probability at least $1-\delta$, $\absns{L_{\cP^{\mathrm{KL}}_{h,s,a}}\Vhat_{h+1} - L_{\cPhat^{\mathrm{KL}}_{h,s,a}}\Vhat_{h+1}} \leq \frac{H}{\rho} \exp{H/\underlambda} \exp{\theta H} \sqrt{{\log{4/(\theta\underlambda\delta)}}/{(2N)}}.$

\end{proposition}
\cref{rem:covering-theta} holds here as well (see \cref{lem:concentration-on-inner-kl}).  
Finally, the proof of \cref{thm:kl-exp-vhat-theorem} is now complete by the application of uniform bound over the $[H]\times\states\times\actions$ space, combined with \cref{eq:main-step-1}.

To prove \cref{thm:kl-pmin-vhat-theorem}, we follow similar steps above. We develop our main concentration inequality result based on \cref{prop:inner-dual-kl}, which now depends on $\pmin$, and the RPVL algorithm property that $\Vhat_{h+1}$ is independent of $\Phat^o_{h,s,a}$ by construction.
\begin{proposition}\label{prop:robust-stochastic-value-est-error-KL-PMIN}
Fix any $h,s,a\in[H]\times\states\times\actions$. For any $\delta\in(0,1)$ and $\rho>0$, we have, with probability at least $1-\delta$,
$\absns{L_{\cP^{\mathrm{KL}}_{h,s,a}}\Vhat_{h+1} - L_{\cPhat^{\mathrm{KL}}_{h,s,a}}\Vhat_{h+1}} \leq  ({H}/{\rho})\sqrt{{\log{2\cardS/\delta}}/{(2N\pmin^2)}}.$

\end{proposition}
We note that the covering number argument is avoided for this result.
Finally, the proof of \cref{thm:kl-pmin-vhat-theorem} is now complete by the application of uniform bound over the $[H]\times\states\times\actions$ space, combined with \cref{eq:main-step-1}.

\paragraph{Wasserstein Uncertainty Set:} We make use of the recent results \citep{gao-2022-distributionally} on  Wasserstein distributionally robust optimization in order to develop the dual reformulation of $L_{\cP^{\mathrm{W}}_{h,s,a}}V$ as given below. 
\begin{proposition}\label{prop:inner-dual-wass}
Consider an RMDP $M$ with the Wasserstein uncertainty set $\cP^{\mathrm{W}}$. Fix any $h,s,a\in[H]\times\states\times\actions$. For any value function $V\in\cV$, the inner optimization problem $L_{\cP^{\mathrm{W}}_{h,s,a}}V$ has the following equivalent form
\begin{align*}
    &L_{\cP^{\mathrm{W}}_{h,s,a}}V = - \inf_{\lambda\in[0,H/\rho^p]} \big(\lambda \rho^p \\&\hspace{1cm}-\bE_{s'\sim P^o_h(\cdot| s,a)}[\inf_{s''\in\states}\curly{V(s'')+\lambda d^p(s'',s')}]\big).
\end{align*}
\end{proposition}
Now we develop our main concentration inequality result based on \cref{prop:inner-dual-wass} and the RPVL algorithm property that $\Vhat_{h+1}$ is independent of $\Phat^o_{h,s,a}$ by construction.
\begin{proposition}\label{prop:robust-stochastic-value-est-error-WASSERSTEIN}
Fix any $h,s,a\in[H]\times\states\times\actions$. For any $\theta,\delta\in(0,1)$ and $\rho>0$, we have, with probability at least $1-\delta$,
\begin{align*}
    &\absns{L_{\cP^{\mathrm{W}}_{h,s,a}}\Vhat_{h+1} - L_{\cPhat^{\mathrm{W}}_{h,s,a}}\Vhat_{h+1}} \leq 2\theta + \\&\hspace{2cm} \frac{H(B_p+\rho^p)}{\rho^p}\sqrt{\frac{\log{\frac{2HB_p+2(H\vee\rho^p)}{\rho^p\theta\delta}}}{2N}}.
\end{align*}
\end{proposition}
\cref{rem:covering-theta} holds here as well (see \cref{lem:cover-number-wasser}).  The proof of \cref{thm:wasserstein-theorem} is now complete by the application of uniform bound over the $[H]\times\states\times\actions$ space, combined with \cref{eq:main-step-1}.

\section{Experiments}

We evaluate the performance of our algorithm on the \textit{Gambler's Problem} environment  \citep[Example 4.3]{sutton2018reinforcement}. This environment is also used in the prior DR-RL works  \citep{zhou2021finite,panaganti2021sample,shi2022distributionally}.

\paragraph{Gambler's Problem:} In the gambler's problem, a gambler starts with a random balance and makes bets on a sequence of coin flips, winning the stake with heads and losing with tails. The game allows $H$ total number of bets. The game also ends when the gambler's balance is either $0$ or $50$. This problem can be formulated as an episodic finite-horizon MDP, with a state space $\states=\curlyns{0,1,\dots,50}$ and an action space $\actions=\curlyns{0,1,\dots,\min\curlyns{s,50-s}}$ at each state $s$. We set the horizon as $H=50$. We denote the heads-up probability as $p_h$. Note that this particular MDP has a stationary transition kernel: $p_h$ remains the same for all $h\in\cH$. The gambler receives a reward of $1$ if $s=50$. The reward is $0$ for all other cases. We use $p_h^o=0.6$ as the nominal model for training the algorithm unless we mention otherwise.

\begin{figure*}[t]
    \centering
    \includegraphics[width=\linewidth]{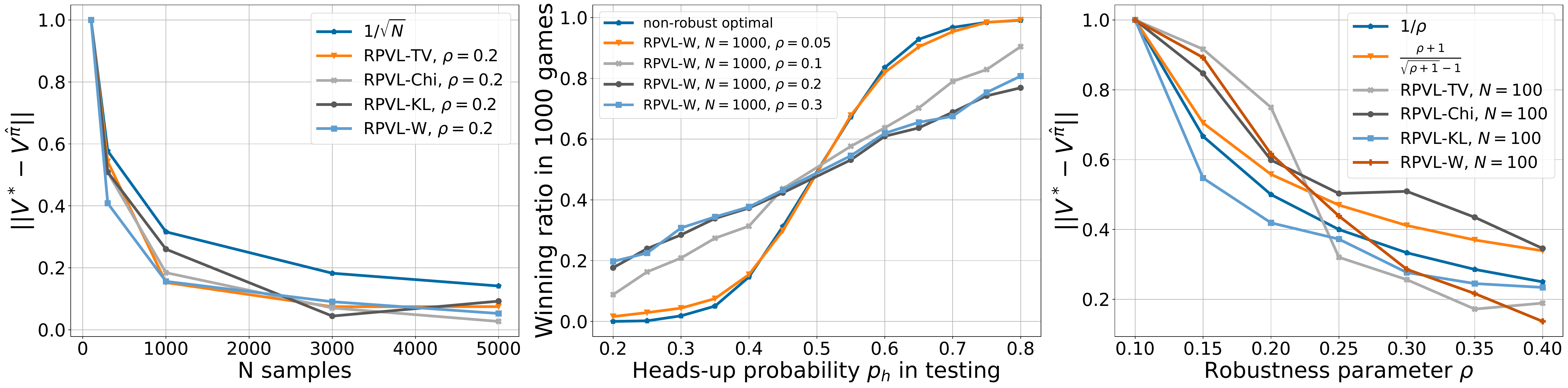}
    \caption{\textit{Convergence of RPVL algorithm.} The left plot shows the rate of convergence with respect to the number of sample $N$. The middle plot shows the level of robustness of Wasserstein robust policies in testing environments with perturbed model parameter $p_h$. The right plot shows how sub-optimality gap changes with respect to the robustness parameter $\rho$.}
    \label{fig:gambler-misc}
\end{figure*}
\begin{figure*}[t]
    \centering 
    \includegraphics[width=\linewidth]{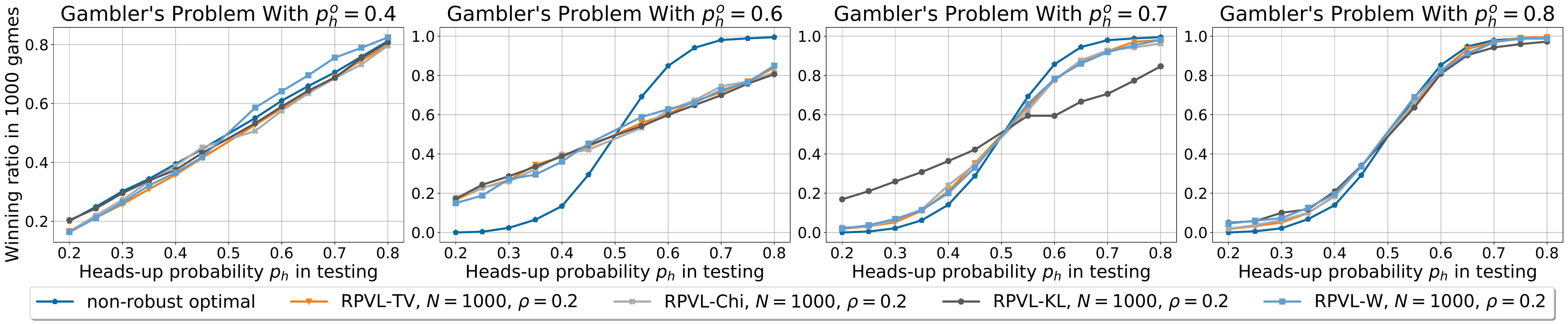}
    \caption{\textit{Robustness fairness.} In each of the four plots, policies are trained with the nominal heads-up probability $p_h^o$. Each plot shows the level of robustness of the policies in perturbed testing environments with different heads-up probability $p_h$.}
    \label{fig:gambler-nominal-study}
\end{figure*}

We study the following characteristics of the RPVL:\\
$(1)$ \textit{Convergence rate of the RPVL algorithm with respect to number of samples $N$.} For each uncertainty set, we train the RPVL policy $\pihat(N)$ with sample sizes $N=100,300,1000,3000,5000$. We demonstrate the convergence by plotting the sub-optimality gap $\normns{V^* - V^{\pihat(N)}}$ against $N$. We also plot $1/\sqrt{N}$ for reference as our results show that  the sub-optimality gap is upper bounded by some function of order $1/\sqrt{N}$ with different coefficients for different uncertainty sets. Hence, we normalize the sub-optimality gap values of each uncertainty set by dividing maximum gap value respectively. As shown in the left plot in Fig. \ref{fig:gambler-misc}, the robust policies of all four uncertainty sets converge to their respective robust optimal policies when $N$ increases.\\
$(2)$ \textit{Effect of robustness parameter $\rho$.} To show this, we consider the Wasserstein uncertainty set. We fix $N=1000$ and train policies with $p_h^o=0.6$ and $\rho=0.05, 0.1,0.2,0.3$. In the middle plot in Fig. \ref{fig:gambler-misc}, it can be seen that the policies trained with lower $\rho$ perform similarly to how non-robust optimal policy performs. As $\rho$ increases, robust policies achieve better winning rate across the perturbed testing environments with $p_h\leq0.5$. When $p_h=0.6$, the testing environment coincides with the training environment, and the optimality of non-robust optimal policy guarantees its optimal performance.\\
$(3)$ \textit{Dependence on $\rho$.} When proving the main theorems, we find that the upper bound of the sub-optimality gap is some function of $\rho$, i.e., of order $1/\rho$ for TV, KL, and Wasserstein and of order $\frac{\rho+1}{\sqrt{\rho+1}-1}$ for Chi-square. We fix $N=100$ and perform policy evaluation on the RPVL policies trained with robustness parameter $\rho$ in $\curlyns{0.1,0.15,\dots,0.4}$. Note that these upper bounds have different coefficients for different uncertainty sets. 
Hence, we normalize the sub-optimality gap values in the same way as we do in (1). As shown in the right plot of Fig. \ref{fig:gambler-misc}, the upper bound for sub-optimality gap is indeed of the order we theorize in the proofs.\\

$(4)$ \textit{Why is robust policy robust?} Our answer follows from the analysis of \textit{Gambler's Problem} in \citep{zhou2021finite}. When the training $p^o_h$ is less than $0.5$, there is a family of optimal policies that achieve a winning ratio that equals to the testing $p_h$, and both the non-robust dynamic programming and RPVL algorithm find such strategy (as shown in the 1st plot in Fig. \ref{fig:gambler-nominal-study}). However, when $p_h^o\geq0.5$, there is a unique non-robust optimal policy: simply bet one dollar at each coin flip. This explains that when the testing $p_h$ increases, the winning ratio increases fast for the non-robust optimal policy, and hence it has an S-shaped curve. On the other hand, for all the robust policies with $\rho=0.2$, the uncertainty sets contain adversarial models with $p_h<0.5$. Thus, all four robust policies still learn the conservative betting strategy (as shown in 2nd plot in Fig. \ref{fig:gambler-nominal-study}). Next, when the training $p_h^o=0.7$, an uncertainty set with radius $\rho=0.2$ may or may not include the adversarial model with $p_h<0.5$ depending on how it measures the distance between probability measures and hence produces policies behaving differently (as shown in 3rd plot in \cref{fig:gambler-nominal-study}). Lastly, in 4th plot in \cref{fig:gambler-nominal-study}, since the uncertainty sets are centered way beyond $p_h=0.5$, all four robust policies converge to the strategy used by non-robust optimal policy. This tells us an important characteristic of robust policies: if uncertainty set is well designed so that it includes meaningful adversarial models, then the corresponding robust policy will manifest robustness.

\begin{remark}
We would like to emphasize that our work is  primarily a theoretical contribution,  addressing  the fundamental sample complexity question of \emph{tabular} (finite state and finite action) distributionally robust RL. So, the goal of this experiments section is mainly to validate the performance our RPVL algorithm in \emph{tabular environment}.   High dimensional  continuous control tasks, such as the MuJoCo environment used in deep RL works, are not tabular and cannot be solved by the tabular methods, including ours. So, we do not consider such continuous control benchmark tasks in this works. We note that the prior works in DR-RL have also considered only tabular environments for experiments \citep{zhou2021finite,panaganti2021sample,shi2022distributionally}.   
\end{remark}

\section{Conclusion}
In this paper, we propose the Robust Phased Value Learning algorithm. Our RPVL algorithm approximates the robust Bellman updates in the standard robust dynamic programming, which is a model-based robust reinforcement learning algorithm. For four distinct uncertainty sets - total variation, chi-square, Kullback-Leibler, and Wasserstein - we present sample complexity results of the learned policy with regard to the optimal robust policy. In order to highlight the theoretical features of the RPVL algorithm, we showcase its performance on the \textit{Gambler's Problem} environment.

This work aims to provide tighter sample complexity results for the finite-horizon robust reinforcement learning problem in the finite state space and action space regime. In comparison to other known prior works, the analyses of our RPVL algorithm provide an improvement in the sample complexity by a factor of $|\states|$ uniformly for all uncertainty sets. We leave the investigation for factor $H$ improvement to our future work.

\section{Acknowledgements}

The authors would like to thank Ruida Zhou for the valuable feedback at an earlier version. This work was supported in part by the National Science Foundation (NSF) grants NSF-CAREER-EPCN-2045783 and NSF ECCS 2038963. Any opinions, findings, and conclusions or recommendations expressed in this material are those of the authors and do not necessarily reflect the views of the sponsoring agencies.


\bibliography{references}
\onecolumn
{\hfill \Large \bf \underline{\Coffeecup ~Supplementary Materials \Coffeecup} \hfill}
\appendix
\section{Technical Results}
\begin{lemma}[Hoeffding's inequality \text{\citep[see][Theorem 2.8]{boucheron2013concentration}}]\label{thm:hoeffding}
 Let $X_1,\dots,X_n$ be independent random variables such that $X_i$ takes its values in $[a_i,b_i]$ almost surely for all $i\leq n$. Let
 \begin{equation*}
     S=\sum_{i=1}^n(X_i-\expect{X_i}).
 \end{equation*}
 Then for every $t>0$,
 \begin{equation*}
     \prob{S\geq t}\leq\exp{-\frac{2t^2}{\sum_{i=1}^n(b_i-a_i)^2}}.
 \end{equation*}
 Furthermore, if $X_1,\dots,X_n$ are a sequence of independent, identically distributed random variables with mean $\mu$. Let $\mean{X}_n = \frac{1}{n}\sum_{i=1}^n X_i$. Suppose that $X_i\in[a,b]$, $\forall i$. Then for all $t>0$
\begin{equation*}
     \prob{\abs{\mean{X}_n - \mu} \geq t} \leq 2\exp{-\frac{2nt^2}{(b-a)^2}}.
\end{equation*}
\end{lemma}

\begin{lemma}[\text{\citealp[Theorem 6.10]{boucheron2013concentration}}]\label{lem:convex-concentration-duchi}
Let $h\colon\bR^n\to\bR$ be convex or concave and $L$-Lipschitz with respect to $\twonorm{\cdot}$. Let $Z_i$ be independent random variables with $Z_i \in [a,b]$. $Z_{1:n}$ denotes the data vector $(Z_1,\dots,Z_n)^T$. For $t\geq 0$,
\begin{equation*}
    \prob{\absns{h(Z_{1:n}) - \expect{h(Z_{1:n})}} \geq t} \leq 2 \exp{-\frac{t^2}{2L^2(b-a)^2}}.
\end{equation*}
\end{lemma}

\begin{lemma}[\text{\citealp[Lemma 7]{duchi2018learning}}]\label{lem:sample-mean-lip-duchi}
The map $\bR^n \ni y \mapsto (\frac{1}{n}\sum_{i=1}^n\absns{y_i}^{k_*})^\frac{1}{k_*}$ is $n^{-\frac{1}{k_* \vee 2}}$-Lipschitz with respect to $\twonorm{\cdot}$.
\end{lemma}

\begin{lemma}[\text{\citealp[Lemma 8]{duchi2018learning}}]\label{lem:expected-sample-mean-concentration-duchi}
Let $k_*\in[1,\infty)$ and let $Y_i$ be an i.i.d. sequence of random variables satisfying $\expect{\absns{Y}^{2k_*}} \leq C^{k_*}\expect{\absns{Y}^{k_*}}$ for some $C\in\bR_+$. For any $k_*\in[1,\infty)$, we have
\begin{equation*}
    \expect{\parent{\frac{1}{n} \sum_{i=1}^n \absns{Y_i}^{k_*}} ^{\frac{1}{k_*}}} \geq \expect{\abs{Y}^{k_*}}^{\frac{1}{k_*}} - \frac{2}{k}\sqrt{C} n^{-\frac{1}{k_* \vee 2}},
\end{equation*}
where $k=k_*/(k_*-1)$ is the conjugate exponent of $k_*$.
\end{lemma}

\subsection{Distributionally Robust Optimization (DRO) Results}
The Total Variation, Chi-square, and Kullback-Liebler uncertainty sets are constructed with the $f$-divergence. The $f$-divergence between the distributions $P$ and $P^o$ is defined as
\begin{equation}\label{eq:f-divergence}
    \infdiv{P}{P^o} = \int f\parent{\frac{dP}{dP^o}} dP^o,
\end{equation}
where $f$ is a convex function \text{\citep{csisz-divergence,moses2011further}}. We obtain different divergences for different forms of the function $f$, including some well-known divergences. For example,  
$f(t) = |t-1|/2$ gives Total Variation, $f(t) = (t-1)^2$ gives chi-square, and $f(t) = t\log t$ gives  Kullback-Liebler.

Let $P^o$ be a distribution on the space $\cX$ and let $l: \cX \to \R$ be a loss function.  We have the following result from the \textit{distributionally robust optimization} literature, see e.g.,~\cite[Proposition 1]{duchi2018learning} and~\cite[Section 3.2]{shapiro2017distributionally}.

\begin{proposition}
\label{prop:dro-inner-problem-solution}
Let $D_{f}$ be the $f$-divergence as defined in \eqref{eq:f-divergence}. Then,
\begin{align}
\label{eq:dro-inner-problem-solution}
   \sup_{D_f(P\|P^o) \leq \rho} \expectsim{P}{l(X)} = \inf_{\lambda> 0, \eta\in\R} ~~ \expectsim{P^o}{\lambda f^* \left(\frac{l(X)-\eta}{\lambda}\right)} + \lambda\rho + \eta,
\end{align}
where $f^*(s) = \sup_{t\geq 0} \{st - f(t)\}$ is the Fenchel conjugate.
\end{proposition}

Note that on the right hand side of \eqref{eq:dro-inner-problem-solution}, the expectation is taken only with respect to $P^{o}$. We will use the above result to derive the dual reformulation of the robust Bellman operator.

\section{Proof of the Theorems}

\subsection{Proof of Theorem 1} \label{appen:tv-vhat-theorem}
Here we provide the proofs of supporting lemmas and that of \cref{thm:tv-vhat-theorem}.

\begin{lemma}[Covering number (TV)]\label{lem:cover-number-tv} Given a value function $V\in\cV$, let $\cU_{V}=\curlyns{(\eta\cdot\vecofone - V)_+\colon \eta\in[0,H]}$. Fix any $\theta \in (0,1)$.  Denote
\begin{equation*}
    \cN_{V}(\theta) = \curly{(\eta\cdot\vecofone - V)_+ \colon \eta\in\curly{\theta, 2\theta, \dots, N_{\theta}\cdot\theta}},
\end{equation*}
where $N_{\theta}= \ceil{H/\theta}$. Then $\cN_{V}(\theta)$ is a $\theta$-cover for $\cU_{V}$ with respect to $\supnorm{\cdot}$, and its cardinality is bounded as $\card{\cN_{V}(\theta)} \leq 2H/\theta$. Furthermore, for any $\nu\in\cN_{V}(\theta)$, we have $\supnorm{\nu}\leq H$.
\end{lemma}
\begin{proof}
First, $N_{\theta}=\ceil{H/\theta}$ is the minimal number of subintervals of length $\theta$ needed to cover $[0,H]$. Denote $J_i = [(i-1)\theta, i\theta)$ to be the $i$-th subinterval, $1\leq i \leq N_{\theta}$. Fix some $\mu\in\cU_{V}.$ Then $\mu = (\eta\cdot\vecofone - V)_+$. Without loss of generality, assume this particular $\eta \in J_i$. Let $\nu = ((i\theta)\cdot\vecofone - V)_+$. Now, for any $s \in \states$,
\begin{align*}
    \abs{\nu(s) - \mu(s)} &= \abs{(i\theta - V(s))_+ - (\eta - V(s))_+} \\
    &\stackleq{(a)} \abs{i\theta - V(s) - \eta + V(s)} \\
    &\leq \abs{i\theta - (i-1)\theta} = \theta,
\end{align*}
where $(a)$ follows from $i\theta>\eta$ and the fact that $\max\curly{x,0}-\max\curly{y,0} \leq x-y$ , if $x > y$.
Taking maximum with respect to $s$ on both sides, we get $\supnorm{\nu - \mu} \leq \theta$. Since $\nu\in \cN_{V}(\theta)$, this suggests $\cN_{V}(\theta)$ is a $\theta$-cover for $\cU_{V}$. The cardinality bound directly follows from
\begin{equation*}
    \card{\cN_{V}(\theta)}=N_{\theta}=\ceil{H/\theta} \leq H/\theta + 1 \leq 2H/\theta,
\end{equation*}
where the last inequality is due to $0<\theta<1\leq H$. Now, for any $\nu\in\cN_{V}(\theta)$, we can establish the following
\begin{equation*}
    \nu = (\eta\cdot\vecofone - V)_+ \leq \parent{H\vecofone - V}_+ \leq H,
\end{equation*}
where the inequality is element-wise.
\end{proof}

\begin{lemma}\label{lem:use-tv-cover}
Fix any $(h,s,a)\in[H]\times\states\times\actions$. Fix any value function $V\in\cV$. Let $\cN_{V}(\theta)$ be the $\theta$-cover of $\cU_{V}=\curlyns{(\eta\cdot\vecofone - V)_+\colon \eta\in[0,H]}$ as described in \cref{lem:cover-number-tv}. We then have
\begin{equation*}
    \sup_{\eta\in[0,H]}\abs{\expectsim{s'\sim \Phat^o_h(\cdot\mid s,a)}{(\eta-V(s'))_+} - \expectsim{s'\sim P^o_h(\cdot\mid s,a)}{(\eta-V(s'))_+}} \leq \max_{\nu\in\cN_{V}(\theta)}\absns{\Phat^o_{h,s,a}\nu - P^o_{h,s,a}\nu} + 2\theta.
\end{equation*}
\end{lemma}
\begin{proof}
For any $\mu\in\cU_{V}$, there exists $\nu\in\cN_{V}(\theta)$ such that $\supnorm{\mu-\nu}\leq\theta$. Now for such particular $\mu$ and $\nu$, we have
\begin{align*}
    \absns{\Phat^o_{h,s,a}\mu - P^o_{h,s,a}\mu} &\leq \absns{\Phat^o_{h,s,a}\mu - \Phat^o_{h,s,a}\nu} + \absns{\Phat^o_{h,s,a}\nu - P^o_{h,s,a}\nu} + \abs{P^o_{h,s,a}\nu - P^o_{h,s,a}\mu} \\
    &\leq \onenormns{\Phat^o_{h,s,a}}\supnorm{\mu-\nu} + \absns{\Phat^o_{h,s,a}\nu - P^o_{h,s,a}\nu} + \onenorm{P^o_{h,s,a}}\supnorm{\nu-\mu} \\
    &\leq \max_{\nu\in\cN_{V}(\theta)}\absns{\Phat^o_{h,s,a}\nu - P^o_{h,s,a}\nu} + 2\theta.
\end{align*}
Taking maximum over $\cU_{V}$ on both sides, we get
\begin{equation*}
    \sup_{\mu\in\cU_{V}}\absns{\Phat^o_{h,s,a}\mu - P^o_{h,s,a}\mu} \leq \max_{\nu\in\cN_{V}(\theta)}\absns{\Phat^o_{h,s,a}\nu - P^o_{h,s,a}\nu} + 2\theta.
\end{equation*}
Now note that by the definition of $\cU_{V}$, we have
\begin{equation*}
    \sup_{\eta\in[0,H]}\abs{\expectsim{s'\sim \Phat^o_h(\cdot\mid s,a)}{(\eta-V(s'))_+} - \expectsim{s'\sim P^o_h(\cdot\mid s,a)}{(\eta-V(s'))_+}} = \sup_{\mu\in\cU_{V}}\absns{\Phat^o_{h,s,a}\mu - P^o_{h,s,a}\mu}.
\end{equation*}
The desired result directly follows.
\end{proof}

\begin{lemma}\label{lem:concentration-on-inner-TV}
Fix any value function $V\in\cV$ and fix $(h,s,a)\in[H]\times\states\times\actions$. For any $\theta,\delta\in(0,1)$ and $\rho>0$, we have, with probability at least $1-\delta$,
\begin{equation*}
    \abs{L_{\cP_{h,s,a}}V - L_{\cPhat_{h,s,a}}V} \leq \sqrt{\frac{H^2\log{4H/\theta\delta}}{2N}} + 2\theta,
\end{equation*}
where $N$ is the number of samples used to approximate $P^o_{h,s,a}$.
\end{lemma}
\begin{proof}
Fix any value function $V\in\cV$ and $(h,s,a)\in[H]\times\states\times\actions$. From \cref{prop:inner-dual-tv}, we have
\begin{align*}
    L_{\cP_{h,s,a}}V &= - \inf_{\eta\in[0,2H/\rho]} \curly{\expectsim{s'\sim P^o_h(\cdot\mid s,a)}{(\eta - V(s'))_+} + \Big(\eta - \inf_{s''\in\states}V(s'')\Big)_+ \cdot \rho - \eta}, \\
    L_{\cPhat_{h,s,a}}V &= - \inf_{\eta\in[0,2H/\rho]} \curly{\expectsim{s'\sim \Phat^o_h(\cdot\mid s,a)}{(\eta - V(s'))_+} + \Big(\eta - \inf_{s''\in\states}V(s'')\Big)_+ \cdot \rho - \eta}.
\end{align*}
Fix any $\rho>0$. Now it follows that
\begin{align}
    \abs{L_{\cP_{h,s,a}}V - L_{\cPhat_{h,s,a}}V} &= \Bigg\lvert \inf_{\eta\in[0,2H/\rho]} \curly{\expectsim{s'\sim \Phat^o_h(\cdot\mid s,a)}{(\eta - V(s'))_+} + \Big(\eta - \inf_{s''\in\states}V(s'')\Big)_+ \cdot \rho - \eta} \nonumber \\
    &\quad- \inf_{\eta\in[0,2H/\rho]} \curly{\expectsim{s'\sim P^o_h(\cdot\mid s,a)}{(\eta - V(s'))_+} + \Big(\eta - \inf_{s''\in\states}V(s'')\Big)_+ \cdot \rho - \eta} \Bigg\rvert \nonumber \\
    &\stackleq{(a)} \sup_{\eta\in[0,2H/\rho]}\abs{\expectsim{s'\sim \Phat^o_h(\cdot\mid s,a)}{(\eta-V(s'))_+} - \expectsim{s'\sim P^o_h(\cdot\mid s,a)}{(\eta-V(s'))_+}} \nonumber \\
    &\leq \max\Bigg\{\sup_{\eta\in[0,H]}\abs{\expectsim{s'\sim \Phat^o_h(\cdot\mid s,a)}{(\eta-V(s'))_+} - \expectsim{s'\sim P^o_h(\cdot\mid s,a)}{(\eta-V(s'))_+}}, \nonumber\\
    &\quad \sup_{\eta\in[H,2H/\rho]}\abs{\expectsim{s'\sim \Phat^o_h(\cdot\mid s,a)}{(\eta-V(s'))_+} - \expectsim{s'\sim P^o_h(\cdot\mid s,a)}{(\eta-V(s'))_+}} \Bigg\} \nonumber\\ 
    &\stackeq{(b)}\max\Bigg\{\sup_{\eta\in[0,H]}\abs{\expectsim{s'\sim \Phat^o_h(\cdot\mid s,a)}{(\eta-V(s'))_+} - \expectsim{s'\sim P^o_h(\cdot\mid s,a)}{(\eta-V(s'))_+}}, \nonumber\\
    &\quad \abs{\expectsim{s'\sim \Phat^o_h(\cdot\mid s,a)}{V(s')} - \expectsim{s'\sim P^o_h(\cdot\mid s,a)}{V(s')}} \Bigg\} \nonumber\\ 
    &\stackleq{(c)} \max\curly{\max_{\nu\in\cN_{V}(\theta)}\absns{\Phat^o_{h,s,a}\nu - P^o_{h,s,a}\nu}+2\theta, \abs{\expectsim{s'\sim \Phat^o_h(\cdot\mid s,a)}{V(s')} - \expectsim{s'\sim P^o_h(\cdot\mid s,a)}{V(s')}}}\label{eq:tv-max-term},
\end{align}
where $(a)$ follows from the fact that $\abs{\inf_x f(x) - \inf_x g(x)} \leq \sup_x\abs{f(x)-g(x)}$. For $(b)$, recall that $\supnormns{V}\leq H$ for any $V\in\cV$. Hence, the term $\eta-V(s')$ is always non-negative for $\eta\in[H,2H/\rho]$, which cancels out by linearity of the expectation. $(c)$ follows from applying \cref{lem:use-tv-cover} to the first term. 
Recall that all $\nu\in\cN_{V}(\theta)$ is upper bounded by $\nu_{\max}:=H$. Now we can apply Hoeffding's inequality (\cref{thm:hoeffding}) to the first term in \cref{eq:tv-max-term}:
\begin{equation*}
    \prob{\absns{\Phat^o_{h,s,a}\nu - P^o_{h,s,a}\nu} \geq \epsilon} \leq 2\exp{-\frac{2N\epsilon^2}{\nu_{\max}^2}} = 2\exp{-\frac{2N\epsilon^2}{H^2}}, \qquad \forall\epsilon>0.
\end{equation*}
Now choose $\epsilon = \sqrt{\frac{H^2\log{2\card{\cN_{V}(\theta)}/\delta}}{2N}}$ and recall that $\card{\cN_{V}(\theta)}\leq 2H/\theta$ from \cref{lem:cover-number-tv}. We have
\begin{align*}
    \prob{\abs{\Phat^o_{h,s,a}\nu - P^o_{h,s,a}\nu} \geq \sqrt{\frac{H^2\log{4H/\theta\delta}}{2N}} } &\leq \prob{\abs{\Phat^o_{h,s,a}\nu - P^o_{h,s,a}\nu} \geq \sqrt{\frac{H^2\log{2\card{\cN_{V}(\theta)}/\delta}}{2N}}} \\
    &\leq \frac{\delta}{\card{\cN_{V}(\theta)}}.
\end{align*}
Applying a union bound over $\cN_{V}(\theta)$, we get
\begin{equation}\label{eq:tv-first-term}
    \max_{\nu\in\cN_{V}(\theta)}\absns{\Phat^o_{h,s,a}\nu - P^o_{h,s,a}\nu} \leq \sqrt{\frac{H^2\log{4H/\theta\delta}}{2N}},
\end{equation}
with probability at least $1-\delta$. Now we can also apply Hoeffding's inequality to the second term in \cref{eq:tv-max-term}. Recall that any value function is bounded by $V_{\max}:=H$. We have
\begin{equation}\label{eq:tv-second-term}
    \absns{\Phat^o_{h,s,a}V - P^o_{h,s,a}V} \leq \sqrt{\frac{H^2\log{2/\delta}}{2N}},
\end{equation}
with probability at least $1-\delta$. Combining \cref{eq:tv-max-term} - \cref{eq:tv-second-term} completes the proof.
\end{proof}

\begin{corollary} \label{cor:random-vec}
Lemma \ref{lem:concentration-on-inner-TV} holds true for any random vector $\Vhat$ independent of $\Phat_{h,s,a}^o$.
\end{corollary}
\begin{proof}
Note that Lemma \ref{lem:concentration-on-inner-TV} holds true for any fixed value function $V$. It is then true for any realization of the random vector $\Vhat$. Now since $\Vhat$ is independent of $\Phat_{h,s,a}^o$, the result directly follows from the law of total probability.
\end{proof}

\begin{proof}[\textbf{Proof of \cref{prop:robust-stochastic-value-est-error-TV}}]
Recall that $\Vhat_{h+1}$ is independent of $\Phat^o_{h,s,a}$ by construction. The result directly follows from Lemma \ref{lem:concentration-on-inner-TV} and Corollary \ref{cor:random-vec}.
\end{proof}

We now have all the ingredients to prove our main result.
\begin{proof}[\textbf{Proof of \cref{thm:tv-vhat-theorem}}]
The proof is inspired by the proof of \text{\cite[Theorem 2.5.1]{kakade-thesis}}. Recall that the optimal robust value function of the RMDP $M = (\states, \actions, \cP, (r_h)_{h=1}^{H}, [H])$ is characterized by a set of functions $\curlyns{V_h^*}_{h=1}^H$. Also, at each time $h$, the RPVL algorithm outputs $\pihat_h$ and $\Vhat_h$. Now,
\begin{equation}\label{eq:decompose-suboptimality-thm1}
    \supnormns{V_1^* - V_1^{\pihat}} \leq \supnormns{V_1^* - \Vhat_1} + \supnormns{\Vhat_1 - V_1^{\pihat}}.
\end{equation}
\textit{Step 1: bounding $V_1^* - \Vhat_1$ in \cref{eq:decompose-suboptimality-thm1}}. For any state $s\in\states$ and time $h\in[H]$, we have
\begin{align*}
    \abs{V_h^*(s) - \Vhat_h(s)} &= \abs{\max_a\curly{r_h(s,a)+L_{\cP_{h,s,a}}V^*_{h+1}} - \max_a\curly{r_h(s,a) + L_{\cPhat_{h,s,a}}\Vhat_{h+1}}} \\
    &\stackleq{(a)} \max_a \abs{L_{\cP_{h,s,a}}V^*_{h+1} - L_{\cPhat_{h,s,a}}\Vhat_{h+1}} \\
    &\leq \max_a\abs{L_{\cP_{h,s,a}}V^*_{h+1} - L_{\cP_{h,s,a}}\Vhat_{h+1}} + \max_a\abs{L_{\cP_{h,s,a}}\Vhat_{h+1} - L_{\cPhat_{h,s,a}}\Vhat_{h+1}} \\
    &\stackleq{(b)}\supnorm{V^*_{h+1} - \Vhat_{h+1}} +  \max_a\abs{L_{\cP_{h,s,a}}\Vhat_{h+1} - L_{\cPhat_{h,s,a}}\Vhat_{h+1}} \\
    &\leq \supnorm{V^*_{h+1} - \Vhat_{h+1}} +  \max_{h,s,a}\abs{L_{\cP_{h,s,a}}\Vhat_{h+1} - L_{\cPhat_{h,s,a}}\Vhat_{h+1}},
\end{align*}
where $(a)$ follows from $\absns{\max_x f(x) - \max g(x)} \leq \max_x\absns{f(x)-g(x)}$. $(b)$ follows from Lemma \ref{lem:uncertainty-inf-lipschitz}. Taking maximum over $\states$, we get
\begin{equation}\label{eq:before-use-recursion-thm1}
    \supnorm{V_h^* - \Vhat_h} \leq \supnorm{V_{h+1}^* - \Vhat_{h+1}} + \max_{h,s,a}\abs{L_{\cP_{h,s,a}}\Vhat_{h+1} - L_{\cPhat_{h,s,a}}\Vhat_{h+1}}.
\end{equation}
Now we use the trivial fact that $V_{H+1}^* = 0 = \Vhat_{H+1}$ and recursively apply \cref{eq:before-use-recursion-thm1} to get
\begin{equation}\label{eq:after-recursion-thm1}
    \supnorm{V_1^* - \Vhat_1} \leq H \max_{h,s,a}\abs{L_{\cP_{h,s,a}}\Vhat_{h+1} - L_{\cPhat_{h,s,a}}\Vhat_{h+1}}.
\end{equation}

\textit{Step 2: bounding $\Vhat_1 - V_1^{\pihat}$ in \cref{eq:decompose-suboptimality-thm1}}. Again, fix any $s\in\states$ and time $h\in[H]$. Note that $\pihat_h$ is the greedy policy with respect to $\Vhat_{h+1}$ as defined in the REVI algorithm. Recall the robust Bellman (consistency) equation (\cref{eq:robust-consistency-eq}). For any deterministic policy $\pi=(\pi_h)_{h\in[H]}$, we have $V_h^\pi(s) = r_h(s,\pi_h(s)) + L_{\cP_{h,s,\pi_h(s)}}V_{h+1}^\pi$. Now using these, we have
\begin{align*}
    \abs{V_h^{\pihat}(s) - \Vhat_h(s)} &= \abs{r_h(s,\pihat_h(s)) + L_{\cP_{h,s,\pihat_h(s)}}V^{\pihat}_{h+1} - r_h(s,\pihat_h(s)) - L_{\cPhat_{h,s,\pihat_h(s)}}\Vhat_{h+1}} \\
    &= \abs{L_{\cP_{h,s,\pihat_h(s)}}V^{\pihat}_{h+1} - L_{\cP_{h,s,\pihat_h(s)}}\Vhat_{h+1} + L_{\cP_{h,s,\pihat_h(s)}}\Vhat_{h+1} - L_{\cPhat_{h,s,\pihat_h(s)}}\Vhat_{h+1}} \\
    &\leq \abs{L_{\cP_{h,s,\pihat_h(s)}}V^{\pihat}_{h+1} - L_{\cP_{h,s,\pihat_h(s)}}\Vhat_{h+1}} + \max_{h,s,a}\abs{L_{\cP_{h,s,a}}\Vhat_{h+1} - L_{\cPhat_{h,s,a}}\Vhat_{h+1}} \\
    &\leq\supnorm{V_{h+1}^{\pihat} - \Vhat_{h+1}} + \max_{h,s,a}\abs{L_{\cP_{h,s,a}}\Vhat_{h+1} - L_{\cPhat_{h,s,a}}\Vhat_{h+1}}.
\end{align*}
Similar to the \textit{step 1}, after taking maximum over $\states$ and unrolling the recursion, we get
\begin{equation}\label{eq:after-recursion-step2-thm1}
    \supnorm{V_1^{\pihat} - \Vhat_1} \leq H \max_{h,s,a}\abs{L_{\cP_{h,s,a}}\Vhat_{h+1} - L_{\cPhat_{h,s,a}}\Vhat_{h+1}}.
\end{equation}

\textit{Step 3: concentration on $\absns{L_{\cP_{h,s,a}}\Vhat_{h+1} - L_{\cPhat_{h,s,a}}\Vhat_{h+1}}$}. Applying \cref{prop:robust-stochastic-value-est-error-TV} and taking a union bound over $(h,s,a)\in[H]\times\states\times\actions$, we get
\begin{equation}\label{eq:concentration-on-vhat-thm1}
    \max_{h,s,a}\abs{L_{\cP_{h,s,a}}\Vhat_{h+1} - L_{\cPhat_{h,s,a}}\Vhat_{h+1}} \leq \sqrt{\frac{H^2\log{4H^2\cardS\cardA/\theta\delta}}{2N}} + 2\theta,
\end{equation}
with probability at least $1-\delta$. Now, plugging \cref{eq:after-recursion-thm1}-\cref{eq:concentration-on-vhat-thm1} into \cref{eq:decompose-suboptimality-thm1}, we get
\begin{equation}
    \supnorm{V_1^* - V_1^{\pihat}} \leq 2H\sqrt{\frac{H^2\log{4H^2\cardS\cardA/\theta\delta}}{2N}} + 4H\theta, \label{eq:N-gap}
\end{equation}
with probability at least $1-\delta$. We can choose $\theta=\epsilon/(8H)$. Note that since $\epsilon\in(0,8H)$, this particular $\theta$ is in $(0,1)$. Now, if we choose
\begin{equation*}
    N \geq \frac{8H^4\log{32H^3\cardS\cardA/\epsilon\delta}}{\epsilon^2},
\end{equation*}
we get $\supnormns{V_1^* - V_1^{\pihat}}\leq\epsilon$ with probability at least $1-\delta$.
\end{proof}

\subsection{Proof of Theorem 2} \label{appen:chi-vhat-theorem}
Here we provide the proofs of supporting lemmas and that of \cref{thm:chi-vhat-theorem}.

\begin{lemma}\label{lem:chi-dro}
Let $D_f$ be defined as in \cref{eq:f-divergence} with the convex function $f(t)=(t-1)^2$ corresponding to the Chi-square uncertainty set. Then
\begin{equation*}
    \inf_{\infdiv{P}{P^o}\leq\rho} \expectsim{P}{l(X)} = - \inf_{\eta\in\bR} \curly{\sqrt{\rho+1}\sqrt{\expectsim{P^o}{(\eta-l(X))^2}} - \eta}.
\end{equation*}
\end{lemma}
\begin{proof}
We first get the Fenchel conjugate of Chi-square divergence. We have \[f^*(s) = \sup_{t\geq 0} (st - (t-1)^2)= \sup_{t\geq 0} ( -t^2 + (s+2)t - 1).\]
It is easy to see that for $s\leq-2$, we have $f^*(s)=-1$. Whenever $s>-2$, from Calculus, the optimal $t$ for this optimization problem is $(s+2)/2$. Hence we get $f^*(s) = \max\curly{s^2/4+s, -1}$. Now, from Proposition \ref{prop:dro-inner-problem-solution} we get, \begin{align*}
    \sup_{D_f(P\|P^o) \leq \rho} \expectsim{P}{l(X)} &= \inf_{\lambda>0, \eta\in\R} \{ \EE_{P^o} [\lambda f^*(\frac{l(X)-\eta}{\lambda})] + \lambda\rho + \eta \} \\
    &= \inf_{\lambda,\eta: \lambda>0, \eta\in\R} \{ \lambda\EE_{P^o} [ \max\{\frac{(l(X)-\eta)^2}{4\lambda^2}+\frac{l(X)-\eta}{\lambda},~-1\}] + \lambda\rho + \eta \}\\
    &= \inf_{\lambda,\eta: \lambda>0, \eta\in\R} \{ \EE_{P^o} [ \max\{\frac{(l(X)-\eta)^2}{4\lambda}+l(X)-\eta,-\lambda\}] + \lambda\rho + \eta \}\\
    &= \inf_{\lambda,\eta: \lambda>0, \eta\in\R} \{ \EE_{P^o} [ (\frac{(l(X)-\eta)^2}{4\lambda}+l(X)-\eta+\lambda)_+ -\lambda] + \lambda\rho + \eta \}\\
    &= \inf_{\lambda,\eta: \lambda>0, \eta\in\R} \{ \frac{1}{4\lambda} \EE_{P^o} [ (l(X)-\eta+2\lambda)^2] + \lambda(\rho-1) + \eta \}\\
    &= \inf_{\lambda,\eta': \lambda>0, \eta'\in\R} \{ \frac{1}{4\lambda} \EE_{P^o} [ (l(X)-\eta')^2] + \lambda(\rho+1) + \eta' \},
\end{align*} where the fourth equality follows form the fact that $\max\{x,y\} = (x-y)_+ + y$ for any $x,y\in\R$, and the last equality follows by making the substitution $\eta'=\eta-2\lambda.$ Taking the optimal value of $\lambda$, i.e.,~$\lambda=\sqrt{\EE_{P^o} [ (l(X)-\eta')^2]}/(2\sqrt{\rho+1})$, we get
\begin{align*}
     \sup_{D_f(P\|P^o) \leq \rho} \EE_{P}[l(X)] = \inf_{ \eta'\in\R} \{ \sqrt{\rho+1}\sqrt{\EE_{P^o} [ (l(X)-\eta')^2]} + \eta' \}.  
\end{align*}
Now,
\begin{align*}
    \inf_{D_f(P\|P^o) \leq \rho} \EE_{P}[l(X)]  &=  -\sup_{D_f(P\|P^o) \leq \rho} \EE_{P}[-l(X)] \\
    &=  -\inf_{ \eta'\in\R} \{ \sqrt{\rho+1}\sqrt{\EE_{P^o} [ (-l(X)-\eta')^2]} + \eta' \}\\
     &=  -\inf_{ \eta\in\R} \{ \sqrt{\rho+1}\sqrt{\EE_{P^o} [ (l(X)-\eta)^2]} - \eta \}, 
\end{align*}
which completes the proof. 
\end{proof}

\begin{proof}[\textbf{Proof of \cref{prop:inner-dual-chi}}]
Applying Lemma \ref{lem:chi-dro} to value function, we get
\begin{equation}
    L_{\cP_{h,s,a}}V = - \inf_{\eta\in\bR} \curly{\sqrt{\rho+1}\sqrt{\expectsim{s'\sim P^o_{h}(\cdot\mid s,a)}{(\eta-V(s'))^2}} - \eta}.
\end{equation}
Now let $h(\eta)=\sqrt{\rho+1}\sqrt{\expectsim{s'\sim P^o_{h}(\cdot\mid s,a)}{(\eta-V(s'))^2}} - \eta$. $h$ is convex in dual variable $\eta$. Now observe that $h(\eta)\geq0$ when $\eta\leq0$ and $h(\frac{\sqrt{\rho+1}}{\sqrt{\rho+1} - 1}H) \geq 0$. Hence it is sufficient to consider $\eta\in[0,\frac{\sqrt{\rho+1}}{\sqrt{\rho+1} - 1}H]$. Setting $C_\rho = \sqrt{\rho + 1}$, we get the desired result.
\end{proof}

\begin{lemma}\label{lem:cover-with-lp-norm}
Fix any $(h,s,a)\in[H]\times\states\times\actions$. Let $\cN_\rho(\theta)$ be the $\theta$-cover of the interval $[0,C_\rho H/(C_\rho -1)]$, where $C_\rho=\sqrt{1+\rho}$. For any value function $V\in\cV$ and $\eta\in[0,C_\rho H/(C_\rho -1)]$, we have
\begin{align*}
    \sup_{\eta\in [0,C_\rho H/(C_\rho -1)]}&\bigg\lvert\sqrt{\expectsim{s'\sim \Phat^o_{h}(\cdot\mid s,a)}{(\eta-V(s'))^2}} - \sqrt{\expectsim{s'\sim P^o_{h}(\cdot\mid s,a)}{(\eta-V(s'))^2}}\bigg\rvert \\
    &\leq \max_{\nu\in\cN_\rho(\theta)}\abs{\sqrt{\expectsim{s'\sim \Phat^o_{h}(\cdot\mid s,a)}{(\nu-V(s'))^2}} - \sqrt{\expectsim{s'\sim P^o_{h}(\cdot\mid s,a)}{(\nu-V(s'))^2}}} + 2\theta.
\end{align*}
Furthermore, we have $\card{\cN_\rho(\theta)}\leq \frac{C_\rho H}{(C_\rho -1)\theta}+ 1$.
\end{lemma}
\begin{proof}
Fix any $\eta\in[0,C_\rho H/(C_\rho -1)]$. Then there exists a $\nu\in\cN_\rho(\theta)$ such that $\absns{\eta-\nu}\leq\theta$. Let $X$ be a random variable that takes values in $\curlyns{V(1),\dots,V(\cardS)}$ with probability $\prob{X=V(s')}=P^o_h(s'\mid s,a)$, for all $s'\in\states$. We use $\normns{\cdot}_{p,P}$ to denote the $L^p$ norm in the measure space on $\states$ possessing a measure determined by the probability mass function $P$. It leads to the $L^p$ norm of a random variable: $\normns{X}_{p,P}=(\bE_P \absns{X}^p)^{1/p}$. Now using these definitions, for the particular $\eta$ and $\nu$ we picked, we have
\begin{align*}
    \bigg\lvert&\sqrt{\expectsim{s'\sim \Phat^o_{h}(\cdot\mid s,a)}{(\eta-V(s'))^2}} - \sqrt{\expectsim{s'\sim P^o_{h}(\cdot\mid s,a)}{(\eta-V(s'))^2}}\bigg\rvert = \abs{\norm{\eta-X}_{2,\Phat^o_{h,s,a}} - \norm{\eta - X}_{2,P^o_{h,s,a}}} \\
    &\leq \abs{\norm{\eta-X}_{2,\Phat^o_{h,s,a}} - \norm{\nu - X}_{2,\Phat^o_{h,s,a}}} + \abs{\norm{\nu-X}_{2,\Phat^o_{h,s,a}} - \norm{\nu - X}_{2,P^o_{h,s,a}}} + \abs{\norm{\nu-X}_{2,P^o_{h,s,a}} - \norm{\eta - X}_{2,P^o_{h,s,a}}} \\
    &\stackleq{(a)} \norm{\eta - X - \nu + X}_{2,\Phat^o_{h,s,a}} + \abs{\norm{\nu-X}_{2,\Phat^o_{h,s,a}} - \norm{\nu - X}_{2,P^o_{h,s,a}}} + \norm{\nu - X - \eta + X}_{2,P^o_{h,s,a}} \\
    &\leq \max_{\nu\in\cN_\rho(\theta)}\abs{\norm{\nu-X}_{2,\Phat^o_{h,s,a}} - \norm{\nu - X}_{2,P^o_{h,s,a}}} + 2\abs{\nu-\eta} \\
    &\leq \max_{\nu\in\cN_\rho(\theta)}\abs{\sqrt{\expectsim{s'\sim \Phat^o_{h}(\cdot\mid s,a)}{(\nu-V(s'))^2}} - \sqrt{\expectsim{s'\sim P^o_{h}(\cdot\mid s,a)}{(\nu-V(s'))^2}}} + 2\theta,
\end{align*}
where $(a)$ is due to reverse triangle inequality. Taking supremum over $\eta\in [0,C_\rho H/(C_\rho -1)]$ on the both sides of the above, we have the desired result. For the cardinality, it is trivial that
\begin{equation}\label{eq:chi-cover-card}
    \card{\cN_\rho(\theta)} \leq \ceil{\frac{C_\rho H}{(C_\rho - 1)\theta}} \leq \frac{C_\rho H}{(C_\rho -1)\theta}+ 1.
\end{equation}
\end{proof}

\begin{lemma}\label{lem:duchi-chi-concentration-result}
Fix any value function $V\in\cV$ and fix $(h,s,a)\in[H]\times\states\times\actions$. We have the following inequality with probability at least $1-\delta$:
\begin{equation*}
    \abs{C_\rho\sqrt{\expectsim{s'\sim \Phat^o_{h}(\cdot\mid s,a)}{(\nu-V(s'))^2}} - C_\rho\sqrt{\expectsim{s'\sim P^o_{h}(\cdot\mid s,a)}{(\nu-V(s'))^2}}} \leq \frac{\sqrt{2}C_\rho^2 H}{(C_\rho-1)\sqrt{N}} \parent{\sqrt{\log{2/\delta}}+1},
\end{equation*}
where $C_\rho = \sqrt{1+\rho}$.
\end{lemma}
\begin{proof}
Fix any $(h,s,a)\in[H]\times\states\times\actions$, $\eta\in[0,C_\rho H/(C_\rho-1)]$, and $V\in\cV$ independent of $P^o_{h,s,a}$. Consider a sequence of i.i.d. samples $\curlyns{X_i}_{i=1}^N$ generated from $P^o_{h,s,a}$. Recall that we use the outcomes of this sequence of random variables to construct $\Phat^o_{h,s,a}$. Besides, we denote $Y_i=(\eta-V(X_i))$. Note that $Y$ is upper bounded by $Y_{\max}:=C_\rho H/(C_\rho-1)$. To ease notation, denote $g(\eta;P)=C_\rho (\expectsim{P}{Y^2})^{1/2}$.

\cref{lem:sample-mean-lip-duchi} implies that $g(\eta;\Phat^o_{h,s,a})=C_\rho(\frac{1}{N}\sum_{i=1}^N Y_i^2)^{1/2}$ is a $\frac{C_\rho}{\sqrt{N}}$-Lipschitz function of the vector $(Y_1,\dots,Y_N)^T$ with respect to $\twonorm{\cdot}$. Applying \cref{lem:convex-concentration-duchi}, we get
\begin{equation}\label{eq:duchi-first-concen-ineq}
    \abs{g(\eta;\Phat^o_{h,s,a}) - \expectsimns{P^o_{h,s,a}}{g(\eta;\Phat^o_{h,s,a})}} \leq \frac{\sqrt{2}C_\rho^2 H}{(C_\rho - 1)\sqrt{N}} \sqrt{\log{2/\delta}},
\end{equation}
with probability at least $1-\delta$. It remains to see that $\expectsimns{P^o_{h,s,a}}{g(\eta;\Phat^o_{h,s,a})}$ and $g(\eta;P^o_{h,s,a})$ are close. Note that the random variable $Y$ satisfies $\expectns{\abs{Y}^{4}}\leq Y_{\max}^{2}\expectns{\abs{Y}^{2}}$. Since $\expectsimns{P^o_{h,s,a}}{(\frac{1}{N}\sum_{i=1}^N Y_i^2)^{1/2}} \leq (\expect{Y^2})^{1/2}$ by Jensen's inequality, \cref{lem:expected-sample-mean-concentration-duchi} implies that
\begin{equation}\label{eq:duchi-second-concen-ineq}
    \abs{\expectsimns{P^o_{h,s,a}}{g(\eta;\Phat^o_{h,s,a})} - g(\eta ; P^o_{h,s,a})} \leq \frac{\sqrt{Y_{\max}}}{\sqrt{N}}= \sqrt{\frac{C_\rho H}{C_\rho-1}}\frac{1}{\sqrt{N}} \leq \frac{C_\rho^2\sqrt{H}}{(C_\rho - 1)\sqrt{N}}.
\end{equation}
Combining \cref{eq:duchi-first-concen-ineq} and \cref{eq:duchi-second-concen-ineq}, we get
\begin{equation*}
    \abs{C_\rho\sqrt{\expectsim{s'\sim \Phat^o_{h}(\cdot\mid s,a)}{(\nu-V(s'))^2}} - C_\rho\sqrt{\expectsim{s'\sim P^o_{h}(\cdot\mid s,a)}{(\nu-V(s'))^2}}} \leq \frac{\sqrt{2}C_\rho^2 H}{(C_\rho-1)\sqrt{N}} \parent{\sqrt{\log{2/\delta}}+1},
\end{equation*}
with probability at least $1-\delta$.
\end{proof}

\begin{lemma}\label{lem:concentration-on-inner-chi}
Fix any value function $V\in\cV$ and $(h,s,a)\in[H]\times\states\times\actions$. For any $\theta,\delta\in(0,1)$ and $\rho > 0$, we have, with probability at least $1-\delta$,
\begin{equation*}
    \abs{L_{\cP_{h,s,a}}V - L_{\cPhat_{h,s,a}}V} \leq \frac{\sqrt{2}C_\rho^2 H}{(C_\rho - 1)\sqrt{N}} \parent{\sqrt{\log{\frac{2(1+C_\rho H/(\theta(C_\rho-1))}{\delta}}} + 1} + 2\theta,
\end{equation*}
where $C_\rho=\sqrt{1+\rho}$ and $N$ is the number of samples used to approximate $P^o_{h,s,a}$.
\end{lemma}
\begin{proof}

From \cref{prop:inner-dual-chi}, we get
\begin{align*}
    L_{\cP_{h,s,a}}V &= - \inf_{\eta\in [0,C_\rho H/(C_\rho -1)]} \curly{C_\rho \sqrt{\expectsim{s'\sim P^o_{h}(\cdot\mid s,a)}{(\eta-V(s'))^2}} - \eta}, \\
    L_{\cPhat_{h,s,a}}V &= - \inf_{\eta\in [0,C_\rho H/(C_\rho -1)]} \curly{C_\rho \sqrt{\expectsim{s'\sim \Phat^o_{h}(\cdot\mid s,a)}{(\eta-V(s'))^2}} - \eta}.
\end{align*}
Now for any $\rho>0$, we have
\begin{align}\label{eq:chi-before-concentration}
    \abs{L_{\cP_{h,s,a}}V - L_{\cPhat_{h,s,a}}V} &= \bigg\lvert \inf_{\eta\in [0,C_\rho H/(C_\rho -1)]} \curly{C_\rho \sqrt{\expectsim{s'\sim \Phat^o_{h}(\cdot\mid s,a)}{(\eta-V(s'))^2}} - \eta} \nonumber \\
    &\quad-\inf_{\eta\in [0,C_\rho H/(C_\rho -1)]} \curly{C_\rho \sqrt{\expectsim{s'\sim P^o_{h}(\cdot\mid s,a)}{(\eta-V(s'))^2}} - \eta} \bigg\rvert \nonumber\\
    &\stackleq{(a)} \sup_{\eta\in [0,C_\rho H/(C_\rho -1)]} \abs{C_\rho\sqrt{\expectsim{s'\sim \Phat^o_{h}(\cdot\mid s,a)}{(\eta-V(s'))^2}} - C_\rho\sqrt{\expectsim{s'\sim P^o_{h}(\cdot\mid s,a)}{(\eta-V(s'))^2}}} \nonumber \\
    &\stackleq{(b)} \max_{\nu\in\cN_\rho(\theta)}\abs{C_\rho\sqrt{\expectsim{s'\sim \Phat^o_{h}(\cdot\mid s,a)}{(\nu-V(s'))^2}} - C_\rho\sqrt{\expectsim{s'\sim P^o_{h}(\cdot\mid s,a)}{(\nu-V(s'))^2}}} + 2\theta,
\end{align}
where $(a)$ is from $\absns{\inf_x f(x) - \inf_x g(x)}\leq \sup_x\absns{f(x)-g(x)}$. $(b)$ is due to \cref{lem:cover-with-lp-norm}. 

Now using \cref{lem:duchi-chi-concentration-result} and applying a union bound over $\nu\in\cN_\rho(\theta)$, we get the following inequalities with probability at least $1-\delta$
\begin{align}\label{eq:chi-final-concentration}
    \max_{\nu\in\cN_\rho(\theta)}\bigg\lvert C_\rho\sqrt{\expectsim{s'\sim \Phat^o_{h}(\cdot\mid s,a)}{(\nu-V(s'))^2}} &- C_\rho\sqrt{\expectsim{s'\sim P^o_{h}(\cdot\mid s,a)}{(\nu-V(s'))^2}} \bigg\rvert \nonumber\\
    &\leq \frac{\sqrt{2}C_\rho^2 H}{(C_\rho - 1)\sqrt{N}}\parent{\sqrt{\log{\frac{2\card{\cN_\rho(\theta)}}{\delta}}} + 1} \nonumber\\
    &\stackleq{(a)} \frac{\sqrt{2}C_\rho^2 H}{(C_\rho - 1)\sqrt{N}}\parent{\sqrt{\log{\frac{2(1+C_\rho H/(\theta(C_\rho-1))}{\delta}}} + 1},
\end{align}
where $(a)$ follows from \cref{eq:chi-cover-card}. Combining \cref{eq:chi-before-concentration} and \cref{eq:chi-final-concentration}, we get the desired inequality.
\end{proof}

\begin{proof}[\textbf{Proof of \cref{prop:robust-stochastic-value-est-error-CHI}}]
Recall that $\Vhat_{h+1}$ is independent of $\Phat^o_{h,s,a}$ by construction. Similar to \cref{cor:random-vec} and \cref{prop:robust-stochastic-value-est-error-TV}, the result directly follows from \cref{lem:concentration-on-inner-chi} and the law of total probability.
\end{proof}

We now have all the ingredients to prove our main result.
\begin{proof}[\textbf{Proof of \cref{thm:chi-vhat-theorem}}]
The proof is almost identical to that of \cref{thm:tv-vhat-theorem}. By applying \cref{prop:robust-stochastic-value-est-error-CHI} and taking a union bound over $(h,s,a)\in[H]\times\states\times\actions$, we have
\begin{equation*}
    \supnorm{V_1^* - V_1^{\pihat}} \leq \frac{2\sqrt{2}C_\rho^2 H^2}{(C_\rho - 1)\sqrt{N}}\parent{\sqrt{\log{\frac{2H\cardS\cardA(1+\frac{C_\rho H}{\theta(C_\rho - 1)})}{\delta}}} + 1} + 4H\theta,
\end{equation*}
with probability at least $1-\delta$. We can choose $\theta=\epsilon/(8H)$. Note that since $\epsilon\in(0,8H)$, this particular $\theta$ is in $(0,1)$. Now, if we choose
\begin{equation*}
    N \geq \frac{32C_\rho^4 H^4}{(C_\rho-1)^2} \frac{1}{\epsilon^2}\parent{\sqrt{\log{\frac{2H\cardS\cardA(1+\frac{8C_\rho H^2}{\epsilon(C_\rho - 1)})}{\delta}}} + 1}^2
\end{equation*}
we get $\supnormns{V_1^* - V_1^{\pihat^*}}\leq\epsilon$ with probability at least $1-\delta$. To simplify the above, consider the elementary inequality $(a+b)^2 \leq 2(a^2+b^2)$, for any non-negative real number $a$ and $b$. With this, we get our final sample complexity result:
\begin{equation*}
    H\cardS\cardA N \geq \frac{64 C_\rho^4}{(C_\rho - 1)^2} \frac{H^5\cardS\cardA}{\epsilon^2}\parent{\log{\frac{2H\cardS\cardA(1+\frac{8C_\rho H^2}{\epsilon(C_\rho - 1)})}{\delta}} + 1}.
\end{equation*}
\end{proof}

\subsection{Proof of Theorem 3} \label{appen:kl-exp-vhat-theorem}
Here we provide the proofs of supporting lemmas and that of \cref{thm:kl-exp-vhat-theorem}.

\begin{lemma}\label{lem:kl-dro}
Let $D_f$ be defined as in \cref{eq:f-divergence} with the convex function $f(t)=t\log{t}$ corresponding to the KL uncertainty set $\cP^{KL}$. Then
\begin{equation*}
    \inf_{\infdiv{P}{P^o}\leq\rho} \expectsim{P}{l(X)} = - \inf_{\lambda \geq 0}\curly{\lambda\rho + \lambda \log{\expectsim{P^o}{\exp{\frac{-l(X)}{\lambda}}}}}.
\end{equation*}
\end{lemma}
\begin{proof}
We first derive the Fenchel conjugate of $f$. We have $f^*(s) = \sup_{t\geq0}\curlyns{st-t\log{t}}$. From calculus, the optimal $t$ for this optimization problem is $\exp{s-1}$. Plugging this back into the conjugate, we get $f^*(s) = \exp{s-1}$. From \cref{prop:dro-inner-problem-solution}, we get
\begin{align*}
    \sup_{D_f(P\|P^o) \leq \rho} \expectsim{P}{l(X)} &= \inf_{\lambda> 0, \eta\in\R} \expectsim{P^o}{\lambda f^* \left(\frac{l(X)-\eta}{\lambda}\right)} + \lambda\rho + \eta \\
    &= \inf_{\lambda\geq0,\eta\in\bR} \curly{\expectsim{P^o}{\lambda\exp{\frac{l(X)-\eta}{\lambda} - 1}} + \lambda\rho + \eta} \\
    &= \inf_{\lambda\geq0} \curly{\lambda\rho + \lambda\log{\expectsim{P^o}{\exp{\frac{l(X)}{\lambda}}}}},
\end{align*}
where the last equality is from infimizing over $\eta$. That is, the optimal $\eta = \lambda\log{\expectsim{P^o}{\exp{\frac{l(X)}{\lambda} - 1}}}$. Now this implies that 
\begin{align*}
    \inf_{D_f(P\|P^o) \leq \rho} \expectsim{P}{l(X)} &= -\sup_{D_f(P\|P^o) \leq \rho} \expectsim{P}{-l(X)} \\
    &= - \inf_{\lambda \geq 0}\curly{\lambda\rho + \lambda \log{\expectsim{P^o}{\exp{\frac{-l(X)}{\lambda}}}}}.
\end{align*}
This completes the proof.
\end{proof}

\begin{proof}[\textbf{Proof of \cref{prop:inner-dual-kl}}]
Applying \cref{lem:kl-dro} to value function, we get
\begin{equation*}
    L_{\cP_{h,s,a}}V = - \inf_{\lambda\geq0} \curly{\lambda\rho + \lambda \log{\expectsim{s'\sim P^o_h(\cdot\mid s,a)}{\exp{\frac{-V(s')}{\lambda}}}}}.
\end{equation*}
We denote $h(\lambda) =\lambda\rho + \lambda \log{\expectsim{s'\sim P^o_h(\cdot\mid s,a)}{\exp{\frac{-V(s')}{\lambda}}}}.$ Note that $h$ is convex in the dual variable $\lambda$. Now fix any $\lambda\geq H/\rho$. We have
\begin{equation*}
    \lambda\rho + \lambda \log{\expectsim{s'\sim P^o_h(\cdot\mid s,a)}{\exp{\frac{-V(s')}{\lambda}}}} \geq \lambda\rho + \lambda \log{{\exp{\frac{-H}{\lambda}}}} \geq 0.
\end{equation*}
In addition, observe that $h$ is monotonically increasing in $\lambda$ when $\lambda\geq H/\rho$. Thus, it is sufficient to optimize $\lambda$ in $[0,H/\rho]$. This implies that
\begin{equation*}
    L_{\cP_{h,s,a}}V = - \inf_{\lambda\in[0,H/\rho]} \curly{\lambda\rho + \lambda \log{\expectsim{s'\sim P^o_h(\cdot\mid s,a)}{\exp{\frac{-V(s')}{\lambda}}}}}.
\end{equation*}
\end{proof}

\begin{lemma}\label{lem:concentration-on-inner-kl}
Fix any value function $V\in\cV$ and $(h,s,a)\in[H]\times\states\times\actions$. For any $\theta,\delta\in(0,1)$ and $\rho>0$, we have, with probability at least $1-\delta$,
\begin{equation*}
    \abs{L_{\cP_{h,s,a}}V - L_{\cPhat_{h,s,a}}V} \leq \frac{H}{\rho} \exp{H/\underlambda} \exp{\theta H} \sqrt{\frac{\log{4/(\theta\underlambda\delta)}}{2N}},
\end{equation*}
where $N$ is the number of samples used to approximate $P^o_{h,s,a}$, and $\underlambda$ is a problem dependent parameter but independent of $N$.
\end{lemma}
\begin{proof}
Fix any value function $V\in\cV$, $(h,s,a)\in[H]\times\states\times\actions$, and any $\rho>0$. From \cref{prop:inner-dual-kl}, we have
\begin{align*}
    L_{\cP_{h,s,a}}V &= - \inf_{\lambda\in [0,H/\rho]} \curly{\lambda\rho + \lambda \log{\expectsim{s'\sim P^o_h(\cdot\mid s,a)}{\exp{\frac{-V(s')}{\lambda}}}}}, \\
    L_{\cPhat_{h,s,a}}V &= - \inf_{\lambda\in [0,H/\rho]} \curly{\lambda\rho + \lambda \log{\expectsim{s'\sim \Phat^o_h(\cdot\mid s,a)}{\exp{\frac{-V(s')}{\lambda}}}}}.
\end{align*}
Suppose the optimization problem $L_{\cP_{h,s,a}}V$ above is achieved at $\lambda^*=0$. It can be shown that when $\lambda^*=0$,  $L_{\cP_{h,s,a}}V=L_{\cPhat_{h,s,a}}V=V_{\min}$ with high probability (and hence $\absns{L_{\cP_{h,s,a}}V-L_{\cPhat_{h,s,a}}V}=0$), where $V_{\min}=\min_{s\in\states}V(s)$, whenever $N$ is greater than some problem dependent constant but independent of the optimality gap $\epsilon$ defined in Theorem \ref{thm:kl-exp-vhat-theorem}. \citep[Appendix C]{nilim-ghaoui-2005} provided a proof of this trivial case in terms of robust optimal control. \citep[Lemma 13]{pmlr-v151-panaganti22a} adapted the proof to RMDP and provided a detailed analysis. We omit the proof here and only discuss the case where $\lambda^*\in(0,H/\rho]$.

Now assume the optimal $\lambda^*$ is achieved in $(0,H/\rho]$. Define $\underlambda=\lambda^*/2$ if $\lambda^* \in (0,1)$ and $\underlambda=1/2$ if $\lambda\geq 1$. We define $\underlambda$ this way because restricting $\underlambda\in(0,1)$ can later give us cleaner expression without compromising its function. Let the optimization problem $L_{\cPhat_{h,s,a}}V$ be achieved at $\hat{\lambda}^*$. Again from \citep[Lemma 4]{zhou2021finite} \citep[Lemma 13]{pmlr-v151-panaganti22a}, it holds that $\hat{\lambda}^*\in(\underlambda,H/\rho]$ whenever $N$ is greater than some problem dependent constant but independent of the optimality gap $\epsilon$ defined in \cref{thm:kl-exp-vhat-theorem}, and hence omit this constant for $N$ further in our analysis.
Now it follows that
\begin{align}
    \abs{L_{\cP_{h,s,a}}V - L_{\cPhat_{h,s,a}}V} &= \Bigg\lvert \inf_{\lambda\in(\underlambda,H/\rho]} \curly{\lambda\rho + \lambda\log{\expectsim{s'\sim \Phat^o_h(\cdot\mid s,a)}{\exp{\frac{-V(s')}{\lambda}}}}} \nonumber \\
    &\quad- \inf_{\lambda\in(\underlambda,H/\rho]} \curly{\lambda\rho + \lambda\log{\expectsim{s'\sim P^o_h(\cdot\mid s,a)}{\exp{\frac{-V(s')}{\lambda}}}}} \nonumber \\
    &\stackleq{(a)} \sup_{\lambda\in(\underlambda,H/\rho]}\abs{\lambda\log{\frac{\expectsim{s'\sim \Phat^o_h(\cdot\mid s,a)}{\exp{\frac{-V(s')}{\lambda}}}}{\expectsim{s'\sim P^o_h(\cdot\mid s,a)}{\exp{\frac{-V(s')}{\lambda}}}}}} \nonumber \\
    &\leq \frac{H}{\rho} \sup_{\lambda\in(\underlambda,H/\rho]}\abs{\log{\frac{\expectsim{s'\sim \Phat^o_h(\cdot\mid s,a)}{\exp{\frac{-V(s')}{\lambda}}} - \expectsim{s'\sim P^o_h(\cdot\mid s,a)}{\exp{\frac{-V(s')}{\lambda}}}}{\expectsim{s'\sim P^o_h(\cdot\mid s,a)}{\exp{\frac{-V(s')}{\lambda}}}}  + 1}} \nonumber \\
    &\stackleq{(b)} \frac{H}{\rho} \sup_{\lambda\in(\underlambda,H/\rho]} \frac{\abs{\expectsim{s'\sim \Phat^o_h(\cdot\mid s,a)}{\exp{\frac{-V(s')}{\lambda}}} - \expectsim{s'\sim P^o_h(\cdot\mid s,a)}{\exp{\frac{-V(s')}{\lambda}}}}}{\abs{\expectsim{s'\sim P^o_h(\cdot\mid s,a)}{\exp{\frac{-V(s')}{\lambda}}}}} \nonumber\\
    &\stackleq{(c)} \frac{H}{\rho} \sup_{\lambda'\in[\rho/H,1/\underlambda)}\frac{\abs{\expectsim{s'\sim \Phat^o_h(\cdot\mid s,a)}{\exp{-\lambda'V(s')}} - \expectsim{s'\sim P^o_h(\cdot\mid s,a)}{\exp{-\lambda'V(s')}}}}{\abs{\expectsim{s'\sim P^o_h(\cdot\mid s,a)}{\exp{-\lambda'V(s')}}}} \label{eq:kl-inner-before-pmin}\\
    &\stackleq{(d)} \frac{H}{\rho}\exp{H/\underlambda} \sup_{\lambda'\in[\rho/H,1/\underlambda)}\abs{\expectsim{s'\sim \Phat^o_h(\cdot\mid s,a)}{\exp{-\lambda'V(s')}} - \expectsim{s'\sim P^o_h(\cdot\mid s,a)}{\exp{-\lambda'V(s')}}} \nonumber\\
    &=\frac{H}{\rho}\exp{H/\underlambda} \sup_{\lambda'\in[\rho/H,1/\underlambda)}\abs{\Phat^o_{h,s,a}\exp{-\lambda'V} - P^o_{h,s,a}\exp{-\lambda'V}} \label{eq:kl-exp-before-concentration},
\end{align}
where $(a)$ follows from the fact that $\abs{\inf_x f(x) - \inf_x g(x)} \leq \sup_x\abs{f(x)-g(x)}$. $(b)$ is due to $\absns{\log{1+x}} \leq \abs{x}$. $(c)$ is from a substitution of $\lambda'=1/\lambda$. $(d)$ is because $\exp{-\lambda'V(s')}$ is lower bounded by $\exp{-H/\underlambda}$, for all $s'\in\states$. In the last equality, $\exp{\cdot}$ denotes the element-wise exponential function.

Now let $\cN_\rho(\theta)$ be a $\theta$-cover of the interval $[\rho/H, 1/\underlambda)$ with $\theta\in(0,1)$. Recall that we defined $\underlambda\in (0,1)$ above. We have $\card{\cN_\rho(\theta)} \leq (1/\underlambda - \rho/H)/\theta + 1 \leq 2/(\theta\underlambda)$. Now fix any $\lambda\in[\rho/H, 1/\underlambda)$, there exists a $\nu\in\cN_\rho(\theta)$ such that $\absns{\lambda-\nu}\leq\theta$. Now for these particular $\lambda$ and $\nu$, we have
\begin{align*}
    \abs{\Phat^o_{h,s,a}\exp{-\lambda V} - P^o_{h,s,a}\exp{-\lambda V}}&=\abs{\sum_{s'\in\states}\parent{\Phat^o_h(s'\mid s,a) - P^o_h(s'\mid s,a)} \exp{-\nu V(s')}\exp{(\nu-\lambda)V(s')}} \\
    &\stackleq{(a)} \abs{\sum_{s'\in\states}\parent{\Phat^o_h(s'\mid s,a) - P^o_h(s'\mid s,a)} \exp{-\nu V(s')}\exp{\theta V_{\max}}} \\
    &= \exp{\theta H}\abs{\Phat^o_{h,s,a}\exp{-\nu V} - P^o_{h,s,a}\exp{-\nu V}} \\
    &\leq \exp{\theta H}\max_{\nu\in\cN_\rho(\theta)}\abs{\Phat^o_{h,s,a}\exp{-\nu V} - P^o_{h,s,a}\exp{-\nu V}},
\end{align*}
where $(a)$ is due to $V_{\max}=H$ and $\abs{\nu-\lambda}\leq\theta$. Taking maximum over $\lambda\in[\rho/H,1/\underlambda)$ on both sides, we get
\begin{equation}
    \sup_{\lambda\in[\rho/H,1/\underlambda)}\abs{\Phat^o_{h,s,a}\exp{-\lambda V} - P^o_{h,s,a}\exp{-\lambda V}} \leq \max_{\nu\in\cN_\rho(\theta)}\abs{\Phat^o_{h,s,a}\exp{-\nu V} - P^o_{h,s,a}\exp{-\nu V}} \exp{\theta H}.
\end{equation}
Observe that for all $\nu\in\cN_\rho(\theta)$ and $s\in\states$, $\exp{-\nu V(s)}\leq \exp{-\nu 0} \leq 1$. That is, $\supnormns{\exp{-\nu V}}\leq 1$. Since $V$ is independent of $\Phat^o_{h,s,a}$, we can apply Hoeffding's inequality (\cref{thm:hoeffding}):
\begin{equation*}
    \prob{\abs{\Phat^o_{h,s,a}\exp{-\nu V} - P^o_{h,s,a}\exp{-\nu V}} \geq \epsilon} \leq 2\exp{-\frac{2N\epsilon^2}{\supnormns{\exp{-\nu V}}^2}} \leq 2\exp{-2N\epsilon^2}, \qquad \forall\epsilon>0.
\end{equation*}
Now choose $\epsilon = \sqrt{\frac{\log{2\card{\cN_\rho(\theta)}/\delta}}{2N}}$. We have
\begin{align*}
    &\prob{\abs{\Phat^o_{h,s,a}\exp{-\nu V} - P^o_{h,s,a}\exp{-\nu V}} \geq \sqrt{\frac{\log{4/(\theta\underlambda\delta)}}{2N}}} \\
    &\qquad\qquad\qquad\qquad\qquad\qquad\leq \prob{\abs{\Phat^o_{h,s,a}\exp{-\nu V} - P^o_{h,s,a}\exp{-\nu V}} \geq \sqrt{\frac{\log{2\card{\cN_\rho(\theta)}/\delta}}{2N}}} \leq \frac{\delta}{\card{\cN_\rho(\theta)}}.
\end{align*}
Applying a union bound over $\nu\in\cN_{\rho,V}(\theta)$, we get
\begin{equation}\label{eq:kl-exp-concentration}
    \max_{\nu\in\cN_\rho(\theta)}\abs{\Phat^o_{h,s,a}\exp{-\nu V} - P^o_{h,s,a}\exp{-\nu V}} \leq \sqrt{\frac{\log{4/(\theta\underlambda\delta)}}{2N}},
\end{equation}
with probability at least $1-\delta$. Combining \cref{eq:kl-exp-before-concentration}-\cref{eq:kl-exp-concentration} completes the proof.
\end{proof}

\begin{proof}[\textbf{Proof of \cref{prop:robust-stochastic-value-est-error-KL-EXP}}]
Recall that $\Vhat_{h+1}$ is independent of $\Phat^o_{h,s,a}$ by construction. Similar to \cref{cor:random-vec} and \cref{prop:robust-stochastic-value-est-error-TV}, the result directly follows from \cref{lem:concentration-on-inner-kl} and the law of total probability.
\end{proof}

We now have all the ingredients to prove our main result.
\begin{proof}[\textbf{Proof of \cref{thm:kl-exp-vhat-theorem}}]
The proof is almost identical to that of \cref{thm:tv-vhat-theorem}. By applying \cref{prop:robust-stochastic-value-est-error-KL-EXP} and taking a union bound over $(h,s,a)\in[H]\times\states\times\actions$, we have
\begin{align*}
    \supnorm{V_1^* - V_1^{\pihat}} &\leq \frac{2H^2}{\rho}\exp{H/\underlambda}\exp{\theta H}\sqrt{\frac{\log{4H\cardS\cardA/(\theta\underlambda\delta)}}{2N}} \\
    &\leq \frac{2H^2}{\rho}\exp{H/\underlambda+H/2}\sqrt{\frac{\log{8H\cardS\cardA/(\underlambda\delta)}}{2N}},
\end{align*}
with probability at least $1-\delta$, and the last inequality is because we can simply choose $\theta=1/2$. Now if we choose
\begin{equation*}
    N \geq \frac{2\exp{3H/\underlambda}}{\rho^2} \frac{H^4\log{8H\cardS\cardA/(\underlambda\delta)}}{\epsilon^2},
\end{equation*}
we get $\supnormns{V_1^*-V_1^{\pihat}}\leq\epsilon$ with probability at least $1-\delta$.
\end{proof}

\subsection{Proof of Theorem 4} \label{appen:kl-pmin-vhat-theorem}
Here we provide the proofs of supporting lemmas and that of \cref{thm:kl-pmin-vhat-theorem}.

\begin{lemma}\label{lem:concentration-on-inner-kl-pmin}
Fix any value function $V\in\cV$ and $(h,s,a)\in[H]\times\states\times\actions$. For any $\theta,\delta\in(0,1)$ and $\rho>0$, we have, with probability at least $1-\delta$,
\begin{equation*}
    \abs{L_{\cP_{h,s,a}}V - L_{\cPhat_{h,s,a}}V} \leq \frac{H}{\rho}\sqrt{\frac{\log{2\cardS/\delta}}{2N\pmin^2}},
\end{equation*}
where $\pmin=\min_{s',s,a,h\colon P_h^o(s'\mid s,a)>0} P^o_h(s'| s,a)>0$ and $N$ is the number of samples used to approximate $P^o_{h,s,a}$.
\end{lemma}
\begin{proof}
From \cref{eq:kl-inner-before-pmin}, we get
\begin{align}
    \abs{L_{\cP_{h,s,a}}V - L_{\cPhat_{h,s,a}}V} &\leq \frac{H}{\rho} \sup_{\lambda\in[\rho/H,1/\underlambda)}\frac{\abs{\expectsim{s'\sim \Phat^o_h(\cdot\mid s,a)}{\exp{-\lambda V(s')}} - \expectsim{s'\sim P^o_h(\cdot\mid s,a)}{\exp{-\lambda V(s')}}}}{\abs{\expectsim{s'\sim P^o_h(\cdot\mid s,a)}{\exp{-\lambda V(s')}}}} \nonumber \\
    &\leq \frac{H}{\rho} \sup_{\lambda\in[\rho/H,1/\underlambda)} \frac{\sum_{s'\in\states} \abs{\Phat^o_h(s'\mid s,a)-P^o_h(s'\mid s,a)} \exp{-\lambda V(s')}}{\sum_{s'\in\states} P^o_h(s'\mid s,a) \exp{-\lambda V(s')}} \nonumber \\
    &\leq \frac{H}{\rho} \sup_{\lambda\in[\rho/H,1/\underlambda)} \frac{\sum_{s'\in\states} \abs{\Phat^o_h(s'\mid s,a)-P^o_h(s'\mid s,a)} \exp{-\lambda V(s')}}{\sum_{s'\in\states} P^o_h(s'\mid s,a) \exp{-\lambda V(s')}} \nonumber \\
    &\stackleq{(a)} \frac{H}{\rho} \max_{s'\colon P^o_h(s'\mid s,a)>0} \abs{\frac{\Phat^o_h(s'\mid s,a)}{P^o_h(s'\mid s,a)} - 1}, \label{eq:kl-pmin-before-concentration}
\end{align}
where $(a)$ follows from $\sum_i a_i/\sum_i b_i\leq \max_i \{a_i/b_i\}$, when $b_i>0$. We denote $\pmin=\min_{s',s,a,h\colon P^o_h(s'\mid s,a)>0} P^o_h(s'\mid s,a)$. Note that $\pmin$ is a problem dependent constant and not dependent on $(h,s,a)$. Now fix any $s'\in\states$ such that $P^o_h(s'\mid s,a)>0$. Hoeffding's inequality tells us
\begin{equation*}
    \prob{\abs{\frac{\Phat^o_h(s'\mid s,a)}{P^o_h(s'\mid s,a)} - 1} \geq \epsilon} \leq 2\exp{-\frac{2N\epsilon^2}{(1/\pmin)^2}} = 2\exp{-2N\pmin^2\epsilon^2}.
\end{equation*}
Choosing $\epsilon=\sqrt{\frac{\log{2\cardS/\delta}}{2N\pmin^2}}$ and applying a union bound over $s'\in\states$, we get
\begin{equation*}
    \max_{s'\colon P^o_h(s'\mid s,a)>0} \abs{\frac{\Phat^o_h(s'\mid s,a)}{P^o_h(s'\mid s,a)} - 1} \leq \sqrt{\frac{\log{2\cardS/\delta}}{2N\pmin^2}},
\end{equation*}
with probability at least $1-\delta$. Combining the above and \cref{eq:kl-pmin-before-concentration}, we get the desired result.
\end{proof}

\begin{proof}[\textbf{Proof of \cref{prop:robust-stochastic-value-est-error-KL-PMIN}}]
Recall that $\Vhat_{h+1}$ is independent of $\Phat^o_{h,s,a}$ by construction. Similar to \cref{cor:random-vec} and \cref{prop:robust-stochastic-value-est-error-TV}, the result directly follows from \cref{lem:concentration-on-inner-kl-pmin} and the law of total probability.
\end{proof}

We now have all the ingredients to prove our main result.

\begin{proof}[\textbf{Proof of \cref{thm:kl-pmin-vhat-theorem}}]
The proof is almost identical to that of \cref{thm:tv-vhat-theorem}. By applying \cref{prop:robust-stochastic-value-est-error-KL-PMIN} and taking a union bound over $(h,s,a)\in[H]\times\states\times\actions$, we have
\begin{equation*}
    \supnorm{V_1^* - V_1^{\pihat}} \leq \frac{2H^2}{\rho} \sqrt{\frac{\log{2H\cardS^2\cardA/\delta}}{2N\pmin^2}},
\end{equation*}
with probability at least $1-\delta$. Now if we choose
\begin{equation*}
    N \geq \frac{2H^4}{\rho^2\pmin^2} \frac{\log{2H\cardS^2\cardA/\delta}}{\epsilon},
\end{equation*}
then we get $\supnormns{V^* - V^{\pihat}} \leq \epsilon$ with probability at least $1-\delta$.
\end{proof}

\subsection{Proof of Theorem 5} \label{appen:wasserstein-theorem}
Here we provide the proofs of supporting lemmas and that of \cref{thm:wasserstein-theorem}.
\begin{proposition}[\text{\citealp[Theorem 1]{gao-2022-distributionally}}]\label{prop:gao-drso-thm}
Let $P^o$ be a distribution on the bounded space $\cX$ and let $l\colon\cX\to\bR$ be a bounded loss function. Then,
\begin{equation*}
    \sup_{\infdivdist{W}{P}{P^o}\leq\rho} \expectsim{P}{l(X)} = \inf_{\lambda\geq0} \expectsim{Y\sim P^o}{\sup_{x\in\cX}\{l(x) - \lambda d^p(x,Y)\}} + \lambda \rho^p.
\end{equation*}
\end{proposition}

\begin{proof}[\textbf{Proof of \cref{prop:inner-dual-wass}}]
Fix any $(h,s,a)\in[H]\times\states\times\actions$. We have
\begin{align*}
    L_{\cP_{h,s,a}}V &= \inf_{\infdivdist{W}{P}{P^o}\leq\rho} \expectsim{s\sim P}{V(s)} = - \sup_{\infdivdist{W}{P}{P^o}\leq\rho} \expectsim{s\sim P}{-V(s)} \\
    &\stackeq{(a)} -\inf_{\lambda\geq0}\curly{\expectsim{s'\sim P^o_h(\cdot\mid s,a)}{\sup_{s''\in\states}\curly{-V(s'')-\lambda d^p(s'',s')}} + \lambda\rho^p} \\
    &= \sup_{\lambda\geq 0}\curly{\expectsim{s'\sim P^o_h(\cdot\mid s,a)}{\inf_{s''\in\states}\curly{V(s'') + \lambda d^p(s'',s')}} - \lambda\rho^p} \\
    &\stackeq{(b)} \sup_{\lambda\in[0,H/\rho^p]}\curly{\expectsim{s'\sim P^o_h(\cdot\mid s,a)}{\inf_{s''\in\states}\curly{V(s'')+\lambda d^p(s'',s')}} - \lambda \rho^p},
\end{align*}
where $(a)$ follows from \cref{prop:gao-drso-thm}. For $(b)$, let us first denote any optimizer in $(a)$ to be $\lambda^*$. Observe that since $V$ is non-negative, it follows that $L_{\cP_{h,s,a}}V$ is also non-negative. Now due to $V_{\max}=H$, we have that
\begin{equation*}
    0 \leq -\lambda^*\rho^p + \expectsim{s'\sim P^o_h(\cdot\mid s,a)}{\inf_{s''}\curly{V(s'')+\lambda d^p(s'',s')}} \leq -\lambda^*\rho^p+\expectsim{s'\sim P^o_h(\cdot\mid s,a)}{V(s')+\lambda d^p(s',s')} \leq -\lambda^*\rho^p+H,
\end{equation*}
where in the last inequality we use that the distance metric satisfies $d(s,s)=0$, for any $s\in\states$.
\end{proof}

\begin{lemma}[Covering number (Wasserstein)]\label{lem:cover-number-wasser}
Fix any value function $V\in\cV$, consider the following set of $\bR^{\cardS}$ vectors:
\begin{equation*}
    \cU_{\rho,V} = \curly{\parent{\inf_{s''\in\states}\curly{V(s'')+\lambda d^p(s'',1)}, \dots, \inf_{s''\in\states}\curly{V(s'')+\lambda d^p(s'',\cardS)}}^T \colon \lambda\in[0,H/\rho^p]}.
\end{equation*}
Let
\begin{equation*}
    \cN_{\rho,V}(\theta) = \curly{\parent{\inf_{s''\in\states}\curly{V(s'')+\lambda d^p(s'',1)}, \dots, \inf_{s''\in\states}\curly{V(s'')+\lambda d^p(s'',\cardS)}}^T \colon \lambda\in \curly{\frac{\theta}{B_p}, \frac{2\theta}{B_p},\dots,N_{\rho,\theta}\frac{\theta}{B_p}}},
\end{equation*}
where $N_{\rho,\theta}=\ceil{\frac{HB_p}{\rho^p\theta}}$ and $B_p=\max_{s',s''}d^p(s'',s')$. Then $\cN_{\rho,V}(\theta)$ is a $\theta$-cover of $\cU_{\rho,V}$ with respect to $\supnorm{\cdot}$, and its cardinality is bounded as $\card{\cN_{\rho,V}(\theta)} \leq \frac{HB_p+(H \vee \rho^p)}{\rho^p\theta}$. Furthermore, for any $\nu\in\cN_{\rho,V}(\theta)$, we have $\supnorm{\nu}\leq \frac{H(B_p+\rho^p)}{\rho^p}$.
\end{lemma}

\begin{proof}
Fix any $\theta\in(0,1)$. First note that $N_{\rho,\theta}$ is the minimal number of subintervals of length $\frac{\theta}{B_p}$ needed to cover $[0,H/\rho^p]$. Denote $J_i= [(i-1)\frac{\theta}{B_p}, i\frac{\theta}{B_p})$, $1\leq i\leq N_{\rho,\theta}$. Fix some $\mu\in\cU_{\rho,V}$. Then $\mu$ must takes the form
\begin{equation*}
    \mu = \parent{\inf_{s''\in\states}\curly{V(s'')+\lambda d^p(s'',1)}, \dots, \inf_{s''\in\states}\curly{V(s'')+\lambda d^p(s'',\cardS)}}^T,
\end{equation*}
for some $\lambda\in[0,H/\rho^p]$. Without loss of generality, assume $\lambda\in J_i$. Now we pick
\begin{equation*}
    \nu = \parent{\inf_{s''\in\states}\curly{V(s'')+ i\frac{\theta}{B_p} d^p(s'',1)}, \dots, \inf_{s''\in\states}\curly{V(s'')+i\frac{\theta}{B_p} d^p(s'',\cardS)}}^T.
\end{equation*}
Fix any $s'\in\states$, we have
\begin{align*}
    \abs{\mu(s') - \nu(s')} &= \abs{\inf_{s''\in\states} \curly{V(s'') + \lambda d^p(s'',s')} - \inf_{s''\in\states} \curly{V(s'') + i\frac{\theta}{B_p} d^p(s'',s)}} \\
    &\stackleq{(a)} \sup_{s''\in\states} \abs{\parent{\lambda - i\frac{\theta}{B_p}} d^p(s'',s')} \\
    &\leq \abs{\lambda - i\frac{\theta}{B_p}} \max_{s',s''} d^p(s'',s') = \abs{\lambda - i\frac{\theta}{B_p}} B_p \\
    &\leq \abs{(i-1)\frac{\theta}{B_p} - i\frac{\theta}{B_p}} B_p = \theta,
\end{align*}
where $(a)$ is due to $\absns{\inf_x f(x)-\inf_x g(x)} \leq \sup_x\absns{f(x)-g(x)}$. Taking maximum over $s'\in\states$ on both sides, we get $\supnorm{\mu-\nu}\leq\theta$. Since $\nu\in\cN_{\rho,V}(\theta)$, this suggests that $\cN_{\rho,V}(\theta)$ is a $\theta$-cover for $\cU_{\rho,V}$.

To bound the cardinality of $\cN_{\rho,V}(\theta)$, we consider two cases. If $0<\rho<1$, then $\rho^p\theta<1$ and
\begin{equation*}
    \ceil{\frac{HB_p}{\rho^p\theta}} \leq \frac{HB_p}{\rho^p\theta} + 1 \leq \frac{HB_p}{\rho^p\theta} + \frac{H}{\rho^p\theta} = \frac{HB_p+H}{\rho^p\theta}.
\end{equation*}
On the other hand, if $\rho>1$, then since $\theta\in(0,1)$, we have
\begin{equation*}
   \ceil{\frac{HB_p}{\rho^p\theta}} \leq \frac{HB_p}{\rho^p\theta} + 1 = \frac{HB_p}{\rho^p\theta} + \frac{\rho^p\theta}{\rho^p\theta} \leq \frac{HB_p}{\rho^p\theta} + \frac{\rho^p}{\rho^p\theta} = \frac{HB_p+\rho^p}{\rho^p\theta}.
\end{equation*}
Hence, we have $\card{\cN_{\rho,V}(\theta)} = N_{\rho,\theta}\leq \frac{HB_p+(H \vee \rho^p)}{\rho^p\theta}$. Now we prove the last claim. Fix any $\nu\in\cN_{\rho,V}$. Note that for any $s'\in\states$,
\begin{equation*}
    \nu(s') = \inf_{s''\in\states}\curly{V(s'') + \lambda d^p(s'',s')} \leq H + \lambda B_p \leq H + \frac{H}{\rho^p}B_p = \frac{H(B_p+\rho^p)}{\rho^p}.
\end{equation*}
The result then follows from taking maximum over $s'\in\states$ on both sides.
\end{proof}

\begin{lemma}\label{lem:use-wasserstein-cover}
Fix any $(h,s,a)\in[H]\times\states\times\actions$. Fix any value function $V\in\cV$. Let $\cN_{\rho,V}(\theta)$ be the $\theta$-cover of the set
\begin{equation*}
    \cU_{\rho,V} = \curly{\parent{\inf_{s''\in\states}\curly{V(s'')+\lambda d^p(s'',1)}, \dots, \curly{V(s'')+\lambda d^p(s'',\cardS)}}^T \colon \lambda\in[0,H/\rho^p]},
\end{equation*}
as described in \cref{lem:cover-number-wasser}. We then have
\begin{align*}
    \sup_{\lambda\in[0,H/\rho^p]} &\abs{\expectsim{s'\sim P^o_h(\cdot\mid s,a)}{\inf_{s''\in\states}\curly{V(s'')+\lambda d^p(s'',s')}} - \expectsim{s'\sim \Phat^o_h(\cdot\mid s,a)}{\inf_{s''\in\states}\curly{V(s'')+\lambda d^p(s'',s')}}} \\
    &\qquad\qquad\qquad\qquad\qquad\qquad\qquad\qquad\qquad\leq \max_{\nu\in\cN_{\rho,V}(\theta)} \abs{\Phat^o_{h,s,a}\nu - P^o_{h,s,a}\nu} + 2\theta.
\end{align*}
\end{lemma}
\begin{proof}
The proof is identical to the proof of \cref{lem:use-tv-cover}.
\end{proof}

\begin{lemma}\label{lem:concentration-on-inner-wasser}
Fix any value function $V\in\cV$ and $(h,s,a)\in[H]\times\states\times\actions$. For any $\theta,\delta\in(0,1)$ and $\rho>0$, we have the following inequality with probability at least $1-\delta$
\begin{equation*}
    \abs{L_{\cP_{h,s,a}}V - L_{\cPhat_{h,s,a}}V} \leq \frac{H(B_p+\rho^p)}{\rho^p}\sqrt{\frac{\log{\frac{2HB_p+2(H\vee\rho^p)}{\rho^p\theta\delta}}}{2N}} + 2\theta,
\end{equation*}
where $B_p=\max_{s',s''}d^p(s'',s')$.
\end{lemma}
\begin{proof}
Fix any value function $V\in\cV$ independent of $\Phat^o_{h,s,a}$ and $(h,s,a)\in[H]\times\states\times\actions$. From \cref{prop:inner-dual-wass}, we have
\begin{align*}
    L_{\cP_{h,s,a}}V &= \sup_{\lambda\in[0,H/\rho^p]}\curly{\expectsim{s'\sim P^o_h(\cdot\mid s,a)}{\inf_{s''\in\states}\curly{V(s'')+\lambda d^p(s'',s')}} - \lambda \rho^p}, \\
    L_{\cPhat_{h,s,a}}V &= \sup_{\lambda\in[0,H/\rho^p]}\curly{\expectsim{s'\sim \Phat^o_h(\cdot\mid s,a)}{\inf_{s''\in\states}\curly{V(s'')+\lambda d^p(s'',s')}} - \lambda \rho^p}.
\end{align*}
Now it follows that
\begin{align}
    \abs{L_{\cP_{h,s,a}}V - L_{\cPhat_{h,s,a}}V}  &= \bigg\lvert \sup_{\lambda\in[0,H/\rho^p]}\curly{\expectsim{s'\sim P^o_h(\cdot\mid s,a)}{\inf_{s''\in\states}\curly{V(s'')+\lambda d^p(s'',s')}} - \lambda \rho^p} \nonumber\\
    &\quad-\sup_{\lambda\in[0,H/\rho^p]}\curly{\expectsim{s'\sim \Phat^o_h(\cdot\mid s,a)}{\inf_{s''\in\states}\curly{V(s'')+\lambda d^p(s'',s')}} - \lambda \rho^p} \bigg\rvert \nonumber\\
    &\stackleq{(a)} \sup_{\lambda\in[0,H/\rho^p]} \bigg\lvert\expectsim{s'\sim P^o_h(\cdot\mid s,a)}{\inf_{s''\in\states}\curly{V(s'')+\lambda d^p(s'',s')}} \nonumber \\
    &\quad- \expectsim{s'\sim \Phat^o_h(\cdot\mid s,a)}{\inf_{s''\in\states}\curly{V(s'')+\lambda d^p(s'',s')}} \bigg\rvert \nonumber\\
    &\stackleq{(b)} \max_{\nu\in\cN_{\rho,V}(\theta)} \abs{\Phat^o_{h,s,a}\nu - P^o_{h,s,a}\nu} + 2\theta, \label{eq:used-wasser-cover}
\end{align}
where $(a)$ follows from $\absns{\sup_x f(x)-\sup_x g(x)} \leq \sup_x\absns{f(x)-g(x)}$. $(b)$ follows from \cref{lem:use-wasserstein-cover}.

Recall that all $\nu\in\cN_{\rho,V}(\theta)$ is bounded by $\nu_{\max}:= \frac{H(B_p+\rho^p)}{\rho^p}$. Now we can apply Hoeffding's inequality (\cref{thm:hoeffding}):
\begin{equation*}
    \prob{\absns{\Phat^o_{h,s,a}\nu - P^o_{h,s,a}\nu} \geq \epsilon} \leq 2\exp{-\frac{2N\epsilon^2}{\nu_{\max}^2}} = 2\exp{-\frac{2N\epsilon^2}{\parent{\frac{H(B_p+\rho^p)}{\rho^p}}^2}}, \qquad \forall\epsilon>0.
\end{equation*}
Now recall that $\card{\cN_{\rho,V}(\theta)} \leq \frac{HB_p+(H\vee\rho^p)}{\rho^p\theta}$ and choose
\begin{equation*}
    \epsilon = \frac{H(B_p+\rho^p)}{\rho^p}\sqrt{\frac{\log{2\card{\cN_{\rho,V}(\theta)}/\delta}}{2N}}.
\end{equation*}
We then have
\begin{align*}
    &\prob{\abs{\Phat^o_{h,s,a}\nu - P^o_{h,s,a}\nu} \geq \frac{H(B_p+\rho^p)}{\rho^p}\sqrt{\frac{\log{\frac{2HB_p+2(H\vee\rho^p)}{\rho^p\theta\delta}}}{2N}}} \\
    &\qquad\qquad\qquad\qquad\leq \prob{\abs{\Phat^o_{h,s,a}\nu - P^o_{h,s,a}\nu} \geq \frac{H(B_p+\rho^p)}{\rho^p}\sqrt{\frac{\log{2\card{\cN_{\rho,V}(\theta)}/\delta}}{2N}}} \\
    &\qquad\qquad\qquad\qquad\leq \frac{\delta}{\card{\cN_{\rho,V}(\theta)}}.
\end{align*}
Finally, applying a union bound over $\cN_{\rho,V}(\theta)$, we get
\begin{equation*}
    \max_{\nu\in\cN_{\rho,V}(\theta)}\absns{\Phat^o_{h,s,a}\nu - P^o_{h,s,a}\nu} \leq \frac{H(B_p+\rho^p)}{\rho^p}\sqrt{\frac{\log{\frac{2HB_p+2(H\vee\rho^p)}{\rho^p\theta\delta}}}{2N}},
\end{equation*}
with probability at least $1-\delta$. Combining the above and \cref{eq:used-wasser-cover} completes the proof.
\end{proof}

\begin{proof}[\textbf{Proof of \cref{prop:robust-stochastic-value-est-error-WASSERSTEIN}}]
Recall that $\Vhat_{h+1}$ is independent of $\Phat^o_{h,s,a}$ by construction. Similar to \cref{cor:random-vec} and \cref{prop:robust-stochastic-value-est-error-TV}, the result directly follows from Lemma \ref{lem:concentration-on-inner-wasser} and the law of total probability.
\end{proof}

We now have all the ingredients to prove our main result.
\begin{proof}[\textbf{Proof of \cref{thm:wasserstein-theorem}}]
The proof is almost identical to that of Theorem \ref{thm:tv-vhat-theorem}. By applying \cref{prop:robust-stochastic-value-est-error-WASSERSTEIN} and taking a union bound over $(h,s,a)\in[H]\times\states\times\actions$, we have
\begin{equation*}
    \supnorm{V_1^* - V_1^{\pihat}} \leq \frac{2H^2(B_p+\rho^p)}{\rho^p}\sqrt{\frac{\log{\frac{2H\cardS\cardA(HB_p+(H\vee\rho^p))}{\rho^p\theta\delta}}}{2N}} + 4H\theta
\end{equation*}
with probability at least $1-\delta$. We can choose $\theta=\epsilon/(8H)$. Note that since $\epsilon\in(0,8H)$, this particular $\theta$ is in $(0,1)$. Now, if we choose
\begin{equation*}
    N \geq \frac{8H^4(B_p+\rho^p)^2}{\rho^{2p}\epsilon^2}\log{\frac{16H^2\cardS\cardA(HB_p+(H\vee\rho^p))}{\rho^p\delta\epsilon}},
\end{equation*}
we get $\supnormns{V_1^*-V_1^{\pihat}}\leq\epsilon$ with probability at least $1-\delta$.
\end{proof}
\end{document}